\documentclass[lettersize,journal]{IEEEtran}
\usepackage{amsmath,amssymb,amsfonts}
\usepackage{algorithm}
\usepackage{algpseudocode}
\usepackage{array}
\usepackage[caption=false,font=normalsize,labelfont=sf,textfont=sf]{subfig}
\usepackage{textcomp}
\usepackage{stfloats}
\usepackage{url}
\usepackage{verbatim}
\usepackage{graphicx}
\usepackage{cite}
\usepackage{xcolor}
\usepackage{colortbl}
\usepackage{booktabs}
\usepackage{multirow}
\usepackage{newfloat}
\usepackage{listings}
\usepackage{arydshln}
\usepackage{caption}

\begin{document}

\title{BEAM: Bi-level Memory-adaptive Algorithmic Evolution for LLM-Powered Heuristic Design}

\author{Chuyang Xiang,
        Yichen Wei,
        Jiale Ma,
        Handing Wang,~\IEEEmembership{Senior Member,~IEEE,}
        Junchi Yan,~\IEEEmembership{Senior Member,~IEEE}
\thanks{Chuyang Xiang and Yichen Wei contributed equally to this work.}}

\markboth{IEEE Transactions on Neural Networks and Learning Systems,~Vol.~XX, No.~X, April~2026}{}

\maketitle

\begin{abstract}
Large Language Model-based Hyper Heuristic (LHH) has recently emerged as an efficient way for automatic heuristic design. However, most existing LHHs just perform well in optimizing a single function within a pre-defined solver. Their  single-layer evolution makes them not effective enough to write a competent complete solver. While some variants incorporate hyperparameter tuning or attempt to generate complex code through iterative local modifications, they still lack a high-level algorithmic modeling, leading to limited exploration efficiency. To address this, we reformulate heuristic design as a Bi-level Optimization problem and propose \textbf{BEAM} (Bi-level Memory-adaptive Algorithmic Evolution). BEAM's exterior layer evolves high-level algorithmic structures with function placeholders through genetic algorithm (GA), while the interior layer realizes these placeholders via Monte Carlo Tree Search (MCTS). We further introduce an Adaptive Memory module to facilitate complex code generation.
To support the evaluation for complex code generation, we point out the limitations of starting LHHs from scratch or from code templates and introduce a Knowledge Augmentation (KA) Pipeline. Experimental results on several optimization problems demonstrate that BEAM significantly outperforms existing LHHs, notably reducing the optimality gap by 37.84\% on aggregate in CVRP hybrid algorithm design. BEAM also designs a heuristic that outperforms SOTA Maximum Independent Set (MIS) solver KaMIS.
\end{abstract}


\section{Introduction}
\label{sec:intro}

Heuristics are crucial for solving complex optimization problems, but manual design is laborious and biased~\cite{MetaDesign}. Automatic Heuristic Design (AHD) emerged to mitigate this issue, with Hyper-Heuristics (HH)~\cite{HHSurvey} automating parameter tuning and components combination—though inflexible. The rise of Large Language Model (LLM)-based code generation~\cite{LLMCode, LLMCoderSurvey} has opened up a new gate for AHD, yet general prompting strategies~\cite{PE, CoT} and general LLM agents fall short for this feedback-intensive task.
 
Language Hyper-Heuristics (LHH) advances AHD by integrating LLM into frameworks such as the Genetic Algorithm (GA)~\cite{ReEvo}. In this line of work, heuristics are treated as individuals, and LLMs are used to improve these individuals iteratively guided by specialized prompts.

However, existing LHHs only perform well in generating a single function of an algorithm~\cite{LLaMEA} instead of entire ones, still demanding manual framework design. This reflects two fundamental limitations of these approaches: \textbf{1)} \textbf{Structural and Prompting Strategy Deficiencies}: Most LHHs are single-layered, treating algorithms as single individuals. When complex requirements are given, their output codes remain simplistic, and the heuristics often fail to evolve after a few generations. While some variants attempt to generate complex code through iterative local modifications~\cite{alphaevolve}, they still lack a high-level algorithmic modeling. These frameworks may also degrade into random search as LLM cannot discern performance causality~\cite{ReEvo} when faced with complex codes. 
\textbf{2)} \textbf{Absent or Deficient Knowledge Augmentation}: Existing approaches either let LLM design algorithms entirely from scratch—providing little or even zero textual external knowledge~\cite{ReEvo} or warm-start them with template functions~\cite{LLM4ADProj} which demands significant manual intervention to design sufficiently diverse templates.
\begin{figure}[tb!]
    \centering
    \includegraphics[width=1\linewidth]{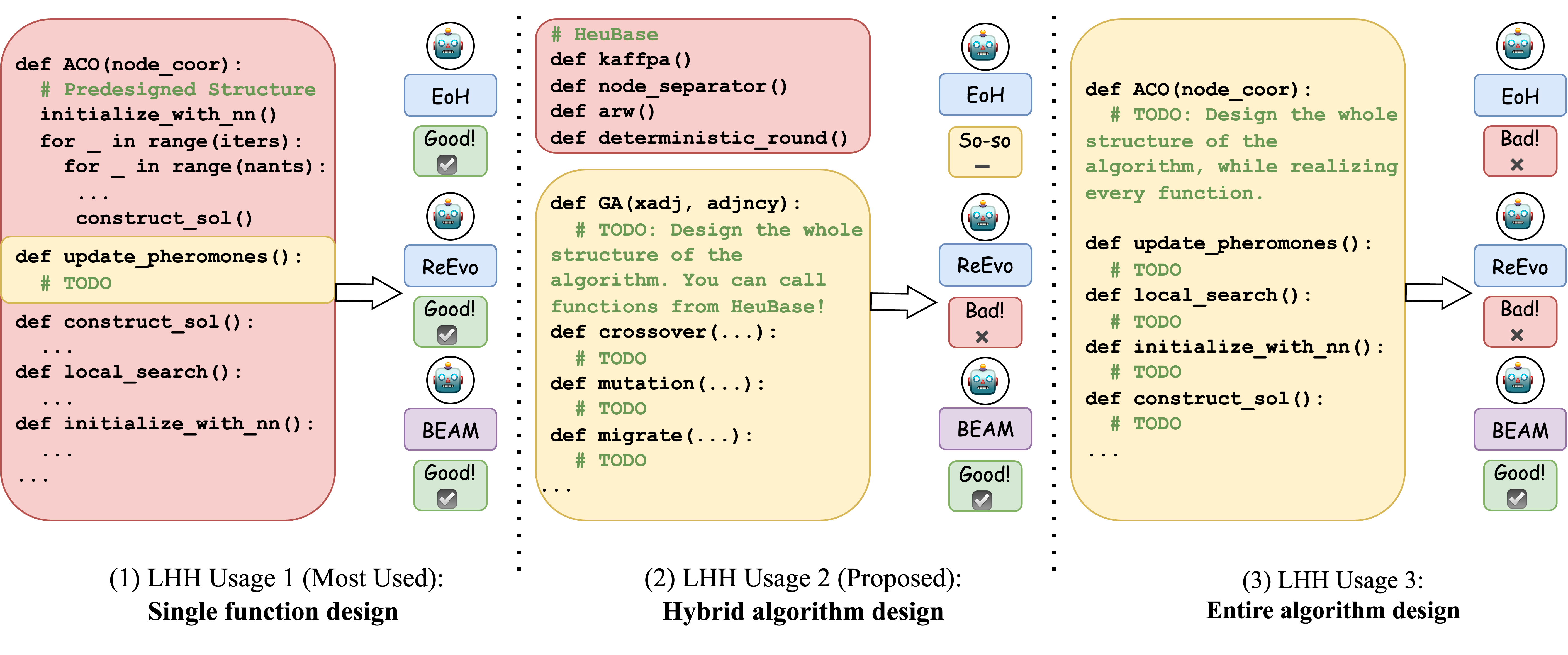}
    \caption{LHH Usage: 1) Single function design (used in~\cite{EoH, ReEvo, PoH}) for a given algorithm structure. 2) Hybrid algorithm design (proposed): designing a whole solver with given heuristic components. 3) Entire algorithm design: designing algorithms from scratch.}
    \label{fig:LHH Usage}
\end{figure}

To overcome these intertwined challenges, we argue that the automated generation of complex heuristics must align more closely with human algorithmic design principles. Human experts rarely construct sophisticated solvers as monolithic entities from scratch; instead, they decompose the problem into high-level structural planning (i.e., the algorithmic framework) and low-level component realization, frequently reusing and recombining established strategies~\cite{KaMIS}. Consequently, we advocate for a paradigm shift in LHHs that reformulates algorithm design as a bi-level optimization problem, allowing specialized search strategies to independently conquer framework evolution and function implementation. Furthermore, to prevent the LLM from conducting blind code exploration, this bi-level search must be firmly grounded in structured external knowledge and a repository of reusable heuristic components. This approach effectively bridges the generative flexibility of modern LLMs with the robust algorithmic recombination principles of traditional HH~\cite{HH}.

To address these limitations and realize this philosophy, we make the following contributions:

\begin{figure*}[tb!]
    \centering
    \includegraphics[width=1\linewidth]{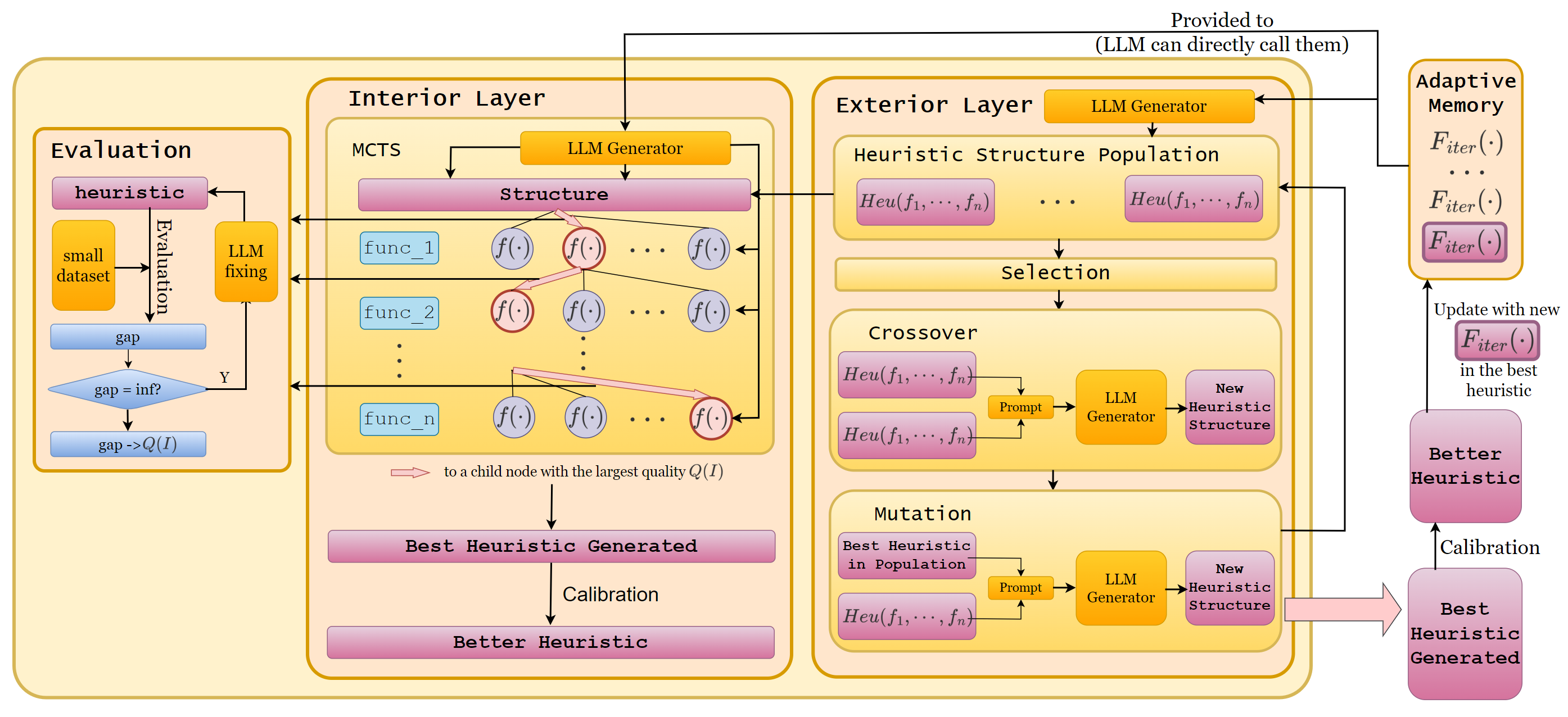}
    \caption{Pipeline of our BEAM. The exterior layer designs the heuristic (code) structure (the heuristic()) via genetic evolution; the interior layer designs the functions (the func\_i()s) by MCTS. Evaluation happens every time when designing a new function to ensure the function quality. After the entire genetic evolution, the best heuristic (code structure + functions) are printed out, and the functions (the func\_i()s) are stored into the adaptive memory for future code structures to call directly. As shown in our experiments (Table~\ref{tab:CombinedTable}), BEAM has better performance over Google's AlphaEvolve~\cite{alphaevolve}.}

    \label{fig:bievo-framework}
\end{figure*}
\begin{itemize}
    \item We propose \textbf{BEAM (Bi-level Memory-adaptive Algorithmic Evolution)} as shown in Section~\ref{sec:bilayer}, which reformulates AHD as a Bi-level Optimization problem ~\cite{Bi-Level}, decomposing it into high-level structure generation (via GA) and low-level function realization (via MCTS). It's further enhanced by an Adaptive Memory mechanism, enabling LLM to directly call elite low-level functions from previous generations.

    \item We introduce a general \textbf{Knowledge Augmentation (KA) pipeline} (in Section~\ref{sec:ka}) where LLM builds 2 datasets: a HeuBase of callable functions and a text-based KnoBase after retrieving external knowledge. We also \textit{construct} part of HeuBase to incorporate cutting-edge heuristic components unavailable via pip, bridging traditional HH idea~\cite{HH} with modern LLM capabilities.

    \item We integrated the KA pipeline to our BEAM and baseline LHHs and tested them on desigining complete solvers for a series of combinatorial and continuous optimization problems. BEAM demonstrates significant performance improvements over existing LHHs and even surpassing SOTA solvers in MIS. In CVRP hybrid algorithm design, it delivers \textbf{37.84\%} aggregate advancement across all benchmarks.

\end{itemize}

\section{Related Work}
\textbf{Prompt Engineering for LLM Coding}
With the rapid progress of LLM in code generation~\cite{LLMCode}, prompt engineering has emerged as a simple yet effective enhancement approach~\cite{PE, LLMCoderSurvey}. Methods like CoT~\cite{CoT} and ToT~\cite{ToT} help structure reasoning, while modular-inspired prompting (e.g. sketch-refine) improves control flow planning~\cite{ChainCoder}.
However, these general-purpose strategies often lack real-time feedback, limiting their effectiveness for black-box optimization tasks where high-quality heuristic generation is crucial~\cite{r1}.

\textbf{LLM For Optimization Problems (LLM4OP).}
While LLM are limited in directly solving complex optimization problems~\cite{optibench}, they excel at problem modeling and code generation. LLM4OP generally falls into two main categories: \textbf{1)} Solver Assistance: LLM translate natural language into formal problem formulations and work with Neural CO solvers~\cite{LNCS}, directly generate solutions~\cite{MCPlanning, optimus} or solver-ready code~\cite{LLMOPT, MA-GTS} using techniques like RAG~\cite{DRoc, RAG} or work with LSTM~\cite{LSTM} to choose algorithms~\cite{AS-LLM}. \textbf{2)} Automatic Heuristic Design (AHD): LLM aid in designing new algorithms or heuristic components. Our proposed framework belongs to the AHD category.

\textbf{Language Hyper Heuristics (LHH).}
Early LHH attempts built a program database and let LLM iteratively refine them~\cite{funsearch}. Later on, researchers were inspired by Hyper Heuristics (HH) and employed LLM as genetic operators~\cite{EC} to evolve new heuristics. Evolution of Heuristics (EoH) used five fixed prompts to do this~\cite{EoH, LLMPromptDesign}, focusing on designing priority functions within predefined frameworks such as Guided Local Search (GLS), a metaheuristic that uses penalty-based guidance to escape local optima. Reflective Evolution (ReEvo) was an improved structure with redesigned prompts and reflection mechanism~\cite{ReEvo}.
LLaMEA introduced a more statistically-sound evaluation method~\cite{LLaMEA}. 
However, population-based search methods~\cite{GA4LLM4AD} often struggle to fully exploit the strengths of individual heuristics~\cite{MCTSLLM}. Recent work integrates RL techniques to mitigate this~\cite{LLMRL4AD, MCTSLLM, PoH}. However, all these LHHs are only good at designing simple functions, suffering from suboptimal framework designs. AlphaEvolve~\cite{alphaevolve} attempts to mitigate this issue by reducing token consumption through LLM-generated modification commands, yet it merely addresses the symptom (token pressure) rather than the root cause (structural limitations). 
Moreover, the conventional evaluation method for LHHs are casual, with reported improvements primarily stem from modifying trivial functions (sometimes even just a simple function returning the maximum of an array~\cite{ReEvo} or a standardized matrix~\cite{PoH} can get the best effects) within suboptimal algorithmic frameworks (often far from SOTA), artificially inflating their perceived capability while offering limited real-world applicability and scientific value.

\textbf{Bi-level Optimization.}
Bi-level Optimization (BLO) is a branch of mathematical programming~\cite{Bi-Level} widely used in the Neural Architecture Search (NAS) field, implemented via evolution-based~\cite{EvoNAS}, gradient-based methods~\cite{DARTS, EasyNAS}.
Nevertheless, BLO applications in HH remain limited due to non-differentiability and large search spaces. Traditional HHs can only be considered as upper-level optimization frameworks that search for effective combinations of metaheuristic and parameter configurations to solve a given optimization problem~\cite{MetaDesign, ACT}. Even in works related to BLO, the outer layer is typically confined to hyperparameter tuning~\cite{GABO4DCP}.

\begin{table*}[t]
\centering
\scriptsize
\caption{Language Hyper Heuristic (LHH) comparison. *: Simple evolution means simply updating all heuristics using different prompts, $\dag$: AlphaEvolve outputs formatted instructions on revising code templates.}
\label{tab:detailedComparison}
\setlength{\tabcolsep}{2.5pt}
\begin{minipage}[t]{\textwidth}
\centering
\begin{tabular}{@{}l p{6.45cm}ll@{}}
\toprule
\textbf{Methods} & \textbf{Individual Strategy} &\textbf{Search Method} &\textbf{Calibration} \\
\midrule
FunSearch~\cite{funsearch} & One-shot & Random sampling & / \\
EoH~\cite{EoH} & One-shot & Simple evolution$^*$ & LLM \\
EvoCAF~\cite{EvoCAF} & One-shot & Simple evolution & LLM \\
HSEvo~\cite{HSEvo} & One-shot & GA & HS \\
LLaMEA-HPO~\cite{LLaMEA-HPO} & One-shot & GA & SMAC3 \\
MCTS-AHD~\cite{MCTSLLM} & One-shot & MCTS & / \\
ReEvo~\cite{ReEvo} & Two-step (text reflection \& codes) & GA & / \\
PoH~\cite{PoH}  & Two-step (text plans \& codes) & MCTS & / \\
CPro1~\cite{CPro1}  & Two-step (text outline \& codes) & Random sampling & Optuna \\
AlphaEvolve~\cite{alphaevolve} & One-shot$^\dag$ & Random sampling & / \\
\textbf{BEAM (Ours)}  & \textbf{Bi-level (algorithm structure \& function realization)} &\textbf{CMA-ES} & \\
\bottomrule
\end{tabular}
\end{minipage}
\begin{minipage}[t]{\textwidth}
\centering
\begin{tabular}{@{}l lp{4.75cm}p{2.5cm}@{}}
\toprule
\textbf{Methods}  & \textbf{KA Type} & \textbf{Benchmark Problems} &\textbf{Type}  \\
\midrule
FunSearch~\cite{funsearch} & Templates  & BPP & Single function \\
EoH~\cite{EoH} & Templates  & BPP, TSP(GLS), FSSP(GLS) & Single function  \\
EvoCAF~\cite{EvoCAF} & / & CAF & Single function \\
HSEvo~\cite{HSEvo} & Text & BPO, TSP(GLS), OP & Single function  \\
LLaMEA-HPO~\cite{LLaMEA-HPO} & / & BBOB & Entire algorithm \\
MCTS-AHD~\cite{MCTSLLM} & / & TSP(GLS, ACO), KP, CVRP(ACO), MKP(ACO), BPP(ACO), CAF  & Single function \\
ReEvo~\cite{ReEvo}  & Templates \& Text & TSP(GLS, ACO, POMO, LEHD), CVRP(ACO, POMO, LEHD), MKP(ACO), BPP(ACO), DPP(GA) & Single function  \\
PoH~\cite{PoH} & / & TSP(GLS), FSSP(GLS) & Single function  \\
CPro1~\cite{CPro1} & / & PA(SA), SymmW(SA), SkewW(SA), BTD(GA), EPA(SA), FR(DFS) & Single function \\
AlphaEvolve~\cite{alphaevolve} & Template \& Text & Some open mathematical construction problems  & Entire algorithm \\
\textbf{BEAM (Ours)} & \textbf{Callable funcs} \& Text & BPP, TSP(GLS), CAF, BBOB, MIS, CVRP, TSP, PMSP  & Entire algorithm \& Hybrid algorithm  \\
\bottomrule
\end{tabular}
\end{minipage}

\end{table*}

\section{BEAM: Bi-level Memory-adaptive Algorithmic Evolution}
\label{sec:bilayer}
As illustrated in Fig.~\ref{fig:bievo-framework}, BEAM is composed of: \textbf{1)} a core bi-layer structure inspired by modular programming~\cite{Modular, ChainCoder} (see Section~\ref{sec:exter} and Section~\ref{sec:inter}); \textbf{2)} an external optimization mechanism called Adaptive Memory (see Section~\ref{sec:am}). We also streamline LHHs and compare them detailedly in Table~\ref{tab:detailedComparison}. All the prompts used are provided in Appendix~\ref{sec:usedPrompts}.

In this paper, instead of treating a complete algorithm as a single entity, we decompose it into a structure and function components to address the challenge of designing complete heuristics. Let \(I\) denote a heuristic individual (a complete algorithm), consisting of an algorithm structure \(\mathcal{S}(I)\) and a set of functions \(\mathcal{F}(I) = \{f_1, \ldots, f_N\}\). 

We measure the \textbf{overall quality} \(Q(I)\) as the average performance (e.g., solution optimality gap or objective value) of individual \(I\) evaluated on a validation set of problem instances. This overall quality decomposes into two components: the \textbf{structure quality} \(Q_s(\mathcal{S}(I))\) represents the inherent effectiveness of the overall algorithmic framework, not including the function implementations, and the \textbf{function quality} \(Q_{f_i}(f_i \mid \mathcal{S}(I))\) measures how well each specific function \(f_i\) implements its role within the given structure (e.g., a neighborhood evaluation function in local search). 

With the abovementioned definition, a bi-level formulation is a must since the quality of a heuristic cannot be determined until all its components are implemented and executed together: $Q(I) = Q_s(\mathcal{S}(I)) + \sum_{i=1}^{N} Q_{f_i}(f_i \mid \mathcal{S}(I))$, where \(N\) is the number of functions required by the structure \(\mathcal{S}(I)\).

Thus, we formulate the bi-level optimization problem as follows:

\begin{align}
    \min_{\alpha}& \quad Q(\alpha, w^*(\alpha)), \label{eq:bilevel_outer}\\
    \text{s.t.}& \quad w^*(\alpha) = \arg\min_{w} Q(\alpha, w), \label{eq:bilevel_inner}
\end{align}

where Eq.~\eqref{eq:bilevel_outer} optimizes the structure, and Eq.~\eqref{eq:bilevel_inner} optimizes function realizations for a given structure. The \textbf{upper-level variable} \(\alpha\) is a symbolic representation (encoded as prompts and code templates for LLM) of the algorithm structure, corresponding to the \textbf{Exterior Layer}. The \textbf{lower-level variable} \(w\) represents the specific function implementations for a given structure \(\alpha\), corresponding to the \textbf{Interior Layer}, and \(w^*(\alpha)\) is the best realization conditioned on \(\alpha\). Note that both \(\alpha\) and \(w\) are discrete symbolic objects in our LLM-based framework, though we use continuous notation for consistency with bi-level optimization literature.

To balance exploration and exploitation, we optimize the Exterior Layer using a Genetic Algorithm (GA) and the Interior Layer using Monte Carlo Tree Search (MCTS) to solve the bi-level problem. The GA evolves the algorithm structures, while MCTS efficiently searches for high-quality function implementations within each structure. Details of the approach are described in the following sections.
We use \(\min\) because \(Q(\cdot)\) represents solution quality measured as optimality gap or cost (lower is better); for maximization problems, objective values are negated before evaluation.

\subsection{Exterior Layer}
\label{sec:exter}
We employ Genetic Evolution~\cite{EC, EvolutionLLM} to evolve heuristic structures, following recent LHH literature~\cite{EoH, ReEvo}. A population at generation \(t\) is \(P^{(t)} = \{I_1^{(t)}, I_2^{(t)}, \ldots, I_n^{(t)}\}\), where each \(I_i^{(t)}\) is a complete heuristic individual with structure \(\mathcal{S}(I_i^{(t)})\) and functions \(\mathcal{F}(I_i^{(t)})\). The population is updated through Algorithm~\ref{alg:exterior}. Individuals are sorted by quality \(Q(I)\) and processed in this quality-ranked order during crossover and mutation operations. Below we describe the evolutionary operators.

\textbf{Population initialization.} This sector initializes $P^{(0)}$ by prompting the LLM with task descriptions, function signatures, requirements,  HeuBase and KnoBase (see Section~\ref{sec:ka}).

\textbf{Education \& Selection.} Before selection, BEAM first educates the population through the \textbf{Education} operation, which sends each structure to the Interior Layer (Section~\ref{sec:inter}) to realize its functions via MCTS, completing the individual. Then, the individuals are sorted according to $Q(I)$ and if the population reaches \texttt{max\_pop\_size}, the worst-performing individuals will be eliminated.

\textbf{Crossover.} For the implementation of $\text{Crossover}(\cdot,\cdot)$ in the prompt level, we simplify ReEvo's approach by: \textbf{1)} eliminating the resource-intensive reflection process~\cite{HSEvo}, instead directly comparing solutions, and \textbf{2)} restricting crossover to algorithm structure only, excluding functions to ensure meaningful comparisons.

\textbf{Mutation.} We also followed ReEvo and used Elitist mutation~\cite{EliMu} on the prompting strategy for $\text{Mutation}(\cdot)$, which requires LLM to learn from the best heuristic individual and redesign the current one. This operator is also performed on algorithm structure.

\begin{algorithm}[H]
\scriptsize
\caption{Genetic Evolutionary Process for Heuristic Individual Structure in BEAM (Exterior Layer). }
\label{alg:exterior}
\begin{algorithmic}[1]
\State $t \gets 0$
\State $P^{(0)} \gets \text{LLM}(\text{task-prompt})$ \Comment{Initialize population via LLM}
\State $P^{(0)} \gets \text{Select}(\text{Education}(P^{(0)})$)) \Comment{Educate and select}
\While{$t < T$} 
    \State Sort $P^{(t)}$ by quality $Q(I)$ in descending order
    \For{$i = 0$ \textbf{ to } $\text{len}(P^{(t)})-2$}
        \If{$\text{Bernoulli}(p_c) = 1$}
            \State $\hat{I}_{\text{cross}} \gets \text{Crossover}(I_i^{(t)}, I_{i+1}^{(t)})$ \Comment{Cross consecutive individuals}
            \State $P^{(t)}_{\text{cross}} \gets P^{(t)}_{\text{cross}} \cup \{\hat{I}_{\text{cross}}\}$
        \EndIf
    \EndFor
    \For{$i = 0$ \textbf{ to } $\text{len}(P^{(t)})-1$}
        \If{$\text{Bernoulli}(p_m) = 1$}
            \State $\hat{I}_{\text{mut}} \gets \text{Mutation}(I_i^{(t)})$ \Comment{Elitist mutation using best individual}
            \State $P^{(t)}_{\text{mut}} \gets P^{(t)}_{\text{mut}} \cup \{\hat{I}_{\text{mut}}\}$
        \EndIf
    \EndFor
    \State $P^{(t+1)} \gets \text{Select}(\text{Education}(P^{(t)} \cup P_{\text{cross}}^{(t)} \cup P_{\text{mut}}^{(t)}))$
    \State $t \gets t + 1$
    \If{$t \bmod \texttt{am\_interval} = 0$} \Comment{Trigger AM every \texttt{am\_interval} generations}
        \State $\text{AM} \gets \text{UpdateAM}(\text{AM}, P^{(t)})$ \Comment{Algorithm~\ref{alg:am}}
    \EndIf
\EndWhile
\State \Return Best individual from $P^{(T)}$
\end{algorithmic}
\end{algorithm}

\subsection{Interior Layer}
\label{sec:inter}

The Interior Layer implements the \textbf{Education Operation}, which completes and evaluates structures proposed by the Exterior Layer. Given a partial structure \(\mathcal{S}(I)\) with placeholders for required functions \(\mathcal{F}(I)=\{f_1, \ldots, f_N\}\), Education realizes these functions, repairs generated code, and calibrates hyperparameters to produce a runnable individual \(I\). The general process is presented in Algorithm~\ref{alg:interior}.

\textbf{Monte-Carlo Tree Search (MCTS).} 

Generally, the MCTS method has the valuation function\footnote{An alternative is One-Shot method, which simultaneously fills in all the functions. The choice between
methods depends on both time constraints and the specific problem requirements.}: 
\(
V(s_t) \leftarrow V(s_t)+\alpha[r_{t+1}+\gamma V(s_{t+1})-V(s_t)]
\). 
Specifically, considering that the number of functions to design is finite and fixed for a certain individual, we use the recursion function:

\begin{equation}
V(f_t) = \max_{f_t} [ r(f_t) +  V_{f_t}(f_{t+1}) ],    
\end{equation}
where
\begin{equation}
\begin{aligned}
    V(f_t) =& Q_s(I) + Q_{f_t}(I) + \cdots + Q_{f_N}(I),\\
    r(f_t)=&Q_{f_t}(I),\\
     V_{f_t}(f_{t+1})=& Q_s(I) + Q_{f_{t+1}}(I) + \cdots + Q_{f_N}(I)
\end{aligned}
\end{equation}
with the \(t\)-th function being \(f_t\).

Note that here we have \(Q_s(I) \in \alpha\) for all individuals \(I\). Therefore, in order to choose the best function set of a heuristic individual, we need to maximize \(V(f_1, \alpha)\), and \(w(\alpha) = \max V(F_1, \alpha)\).

In practice, for each function in the given structure, we try several different realization of the function, and then fill in all the functions that remains unrealized. After the evaluation of the entire structure, we select the best one's function realization. This process will loop until all the functions are properly realized.

\textbf{Fixing.} We add a fixing process~\cite{LLMDebug} after the function fill-in process following LLaMEA~\cite{LLaMEA} to address code errors. BEAM specifically handles: \textbf{1)} compile/runtime errors, and \textbf{2)} constraint violations, ensuring full heuristic exploitation. 

\textbf{Calibration.}    
Following prior work~\cite{HSEvo, CPro1}, we add a calibration (hyperparameter tuning) feature: we require LLM to give a hyperparameter test range and then utilize the traditional technique CMA-ES~\cite{CMA-ES} for calibration. 

\begin{algorithm}[H]
\scriptsize
\caption{Monte-Carlo Tree Search with Fixing and Calibration for Function Realization in BEAM (Interior Layer).}
\label{alg:interior}
\begin{algorithmic}[1]
\For{$t = 1$ \textbf{ to } $N$} \Comment{$N$ = total number of functions}
    \State $\texttt{FuncList} \gets \text{LLM}(I, \text{Fill-}t\text{-prompt})$ \Comment{Generate $m$ candidates for $f_t$}
    \State $I^* \gets \text{LLM}(I, \texttt{FuncList}_0, \text{Fill-all-prompt})$ \Comment{Complete with $f_t = \texttt{FuncList}_0$}
    \State $I^* \gets \text{Fix}(I^*)$
    \State $best\_idx \gets 0$
    \For{$j = 1$ \textbf{ to } $m-1$} \Comment{Try remaining candidates}
        \State $I^{\text{temp}} \gets \text{LLM}(I, \texttt{FuncList}_j, \text{Fill-all-prompt})$
        \State $I^{\text{temp}} \gets \text{Fix}(I^{\text{temp}})$
        \If{$Q(I^{\text{temp}}) > Q(I^*)$} \Comment{Higher quality is better}
            \State $I^* \gets I^{\text{temp}}$ \Comment{Update best individual}
            \State $best\_idx \gets j$
        \EndIf
    \EndFor
    \State $I \gets I$ with $f_t \gets \texttt{FuncList}_{best\_idx}$ \Comment{Fix $f_t$ to best candidate}
\EndFor
\State $I \gets \text{Calibration}(I)$ \Comment{Tune hyperparameters}
\State \Return $I$
\end{algorithmic}
\end{algorithm}

\subsection{Adaptive Memory}
\label{sec:am}
We also introduce a mechanism called Adaptive Memory (AM) to facilitate complex code generation. AM happens when the evolutionary process needs to reset population~\cite{Restart}. We present a new way beyond simply retaining elite individuals while injecting random new members~\cite{Restart2}.

The motivation behind AM is to make previously generated, high-quality functions directly reusable by the LLM. In LLM-based heuristic design, repeatedly generating long code blocks for every new candidate is costly and noisy. Instead of reducing output size by `patching' existing code like AlphaEvolve~\cite{alphaevolve}, BEAM lets the LLM recall and call stored functions via importing them, greatly shrinking the amount of code it must emit on each generation.

This approach also promotes diversity: by exposing a pool of reusable components, AM encourages the LLM to combine them in new ways just like how innovation often happens in scientific research.

In practice, AM selects functions generated through MCTS that appear in elite solutions at the end of every \texttt{am\_interval} generations (see Algorithm~\ref{alg:exterior}), and updates the pool iteratively by inserting strong new functions and retiring outdated ones. AM only provides the LLM with the stored functions' names and purpose statements, allowing the LLM to directly import them.

For each candidate function \(f\in\mathcal{C}\), we compute a composite score:
\begin{equation}
  S(f) = \alpha_1 \tilde{F}_{\text{fit}}(f) + \alpha_2 \tilde{F}_{\text{nov}}(f) + \alpha_3 \tilde{F}_{\text{use}}(f) - \alpha_4 \tilde{F}_{\text{age}}(f),
  \label{eq:am_score}
\end{equation}
with novelty measured by $F_{\text{nov}} = 1-\max_{g\in \text{AM}}\text{sim}(f,g)$. If \(f\) is very similar to an existing memory entry \(g^*\) (similarity $>\tau$), it only replaces \(g^*\) when its score exceeds \(S(g^*) + \Delta_{\text{th}}\).

AM also maintains a fixed capacity \(C_{\max}\) and evicts low-utility entries using:
\begin{equation}
  U^*(f) = \lambda S(f) + (1-\lambda) \bar{\Delta}(f),
  \label{eq:long_term_utility}
\end{equation}
where \(\bar{\Delta}(f)\) is the exponential moving average of recent fitness improvements. When capacity is exceeded, the lowest-utility functions are removed, and rarely-used entries are pruned periodically.

The algorithm is shown in Algorithm~\ref{alg:am}. Key notation: $\mathcal{C}$ is the candidate set from top elites, $S(f)$ is the selection score, and $U^*(f)$ is the long-term utility used for eviction decisions.

\begin{algorithm}[H]
\caption{Adaptive Memory insertion and maintenance.}
\scriptsize
\label{alg:am}
\begin{algorithmic}[1]
\State $\mathcal{C} \gets$ extract functions from top-$E$ elite individuals
\For{each $f \in \mathcal{C}$}
  \State Compute $S(f)$ using Eq.~\eqref{eq:am_score}
  \State Find most similar function: $g^* = \arg\max_{g\in \text{AM}} \text{sim}(f,g)$
  \If{$\text{sim}(f, g^*) > \tau$} \Comment{High similarity detected}
    \If{$S(f) > S(g^*) + \Delta_{\text{th}}$} 
      \State Replace $g^*$ with $f$ in AM
    \Else 
      \State Discard $f$ \Comment{Not sufficiently better}
    \EndIf
  \Else
    \State Insert $f$ into AM \Comment{Novel function}
  \EndIf
\EndFor
\While{$|\text{AM}| > C_{\max}$} \Comment{Capacity overflow}
  \State Evict function with smallest $U^*(f)$ (Eq.~\eqref{eq:long_term_utility})
\EndWhile
\State Remove functions unused for $T_{\text{idle}}$ generations with $U^*(f) < \varepsilon$
\end{algorithmic}
\end{algorithm}

Note that this mechanism is uniquely enabled by our framework: traditional LHHs cannot support it due to their single-layer structure.
 
In all, AM allows LLM to retain high-performing functions from previous generations without realizing them again while encouraging new combinations.

\section{Knowledge Augmentation Pipeline for LHH Evaluation}

This section presents our Knowledge Augmentation (KA) pipeline for evaluating LHHs on complex code generation and complete solver design. We first discuss limitations in existing LHH evaluation practices (Section~\ref{sec:tbl}), then introduce our KA pipeline and its intended evaluation focus (Section~\ref{sec:ka}).

\subsection{Limitations of Existing LHH Evaluation} 
\label{sec:tbl}

Current LHH benchmarks in the optimization field suffer from a fundamental mismatch with AHD requirements: \textbf{1)} Most of them focus on designing small functions within predefined algorithms for problems like CO~\cite{EoH, PoH}. They strongly rely on the human-designed external solver frameworks to achieve good results. \textbf{2)} Most of them evaluated LHH-designed individual functions within suboptimal CO solvers~\cite{ReEvo, HSEvo}. Reported improvements often stem from modifying trivial functions—in some cases, even a simple function computing array maxima~\cite{ReEvo} or normalizing matrices~\cite{PoH} is sufficient to achieve SOTA results. Consequently, these benchmarks offer limited practical utility. 
Moreover, the metric is also arbitrarily defined. Details can be found in Table~\ref{tab:detailedComparison}.

\subsection{KA-Guided Evaluation} 
\label{sec:ka}
The need for KA comes from two observations about LLMs in heuristic design:
\begin{itemize}
    \item LLMs struggle to create very complex heuristics from scratch; they often either \textbf{get stuck at initial solutions} or \textbf{generate constraint-violating outputs}.
    \item Yet LLMs are strong at \textbf{repurposing components} and at \textbf{designing meta-frameworks} such as outer architectures, parameter schedules, and heuristic coordination.
\end{itemize}

Therefore, the core idea is to let LHHs design complete solvers: it evaluates how well an LHH composes reusable components, integrates domain knowledge, and produces working algorithms without relying on externally fixed solver frameworks.

\begin{figure}[tb!]
    \centering
    \includegraphics[width=0.85\linewidth]{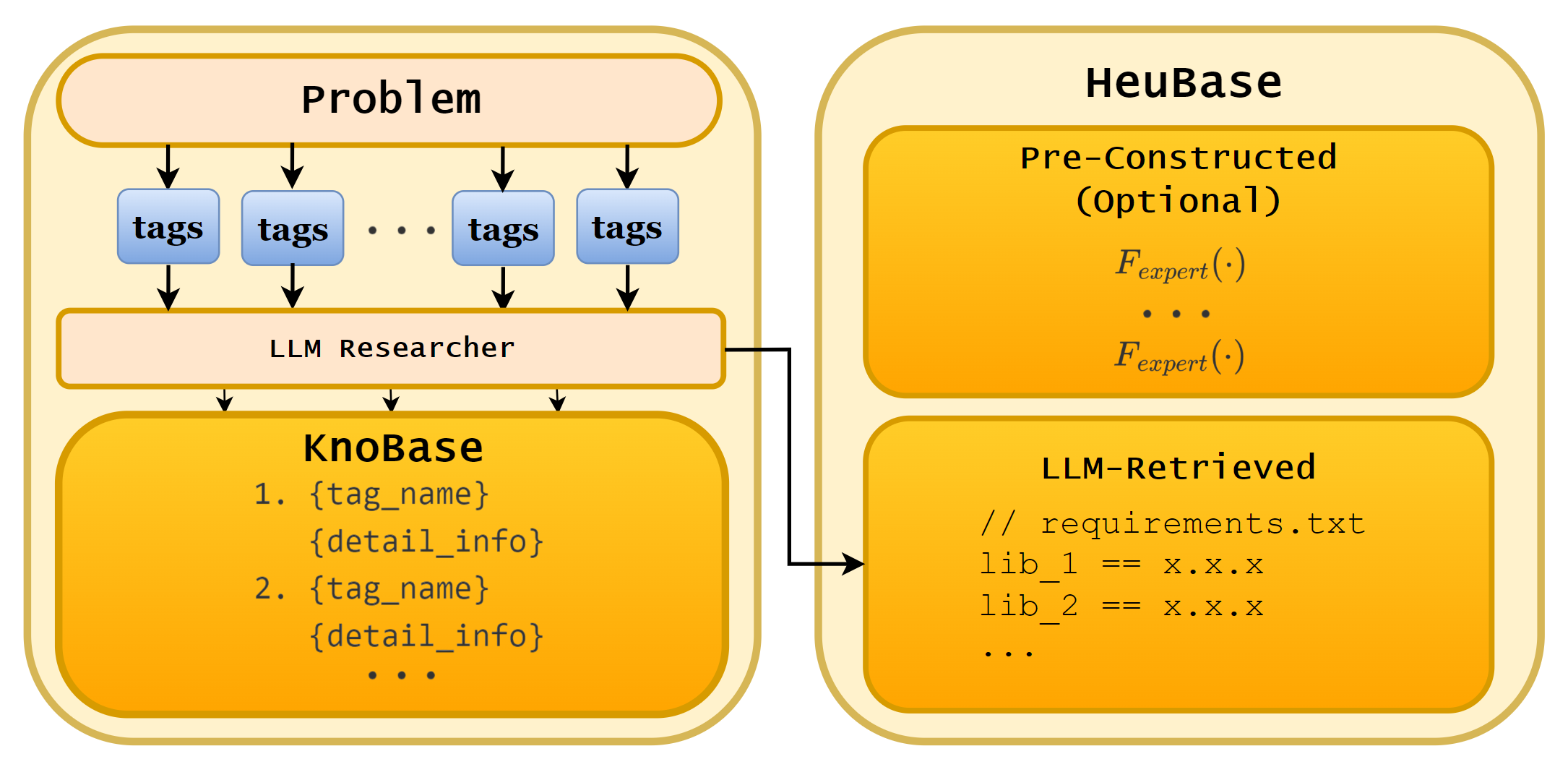}
    \caption{Proposed KA pipeline. For each task, LLMs first summarize problem-specific tags and then use them to construct KnoBase and the LLM-retrieved part of HeuBase.}
    \label{fig:pkr}
\end{figure}

With this idea, Fig.~\ref{fig:pkr} shows our KA pipeline, where LLMs build two structured databases: a \textbf{HeuBase} of callable functions and a text-based \textbf{KnoBase} of retrieved prior knowledge.

\paragraph{HeuBase}
HeuBase is a reusable heuristic component repository that LLMs can directly call, unlike AlphaEvolve's template-based Program Database~\cite{funsearch, alphaevolve}. We provide only function names and brief descriptions, similar to AM, and show that this supports novel combinations while preserving diversity (See~\ref{sec:diversityDiscussions}).

HeuBase has two parts: \textbf{LLM-Retrieved} components, which are pip-installable libraries identified by LLM researchers from task tags and added via \texttt{requirements.txt}; and \textbf{Pre-Constructed} components, which are handcrafted heuristic routines (e.g. optimized 2-opt variants~\cite{2OPT}) unavailable from pip packages. The database is intentionally compact but designed to grow through community contribution.

Notably, while our KA pipeline focuses on designing complete solvers by composing reusable components, it establishes a complementary relationship with existing single-function LHHs. These LHHs excel at iteratively refining individual functions within fixed algorithmic frameworks, and their optimized single functions can be seamlessly integrated into HeuBase, enriching the repository and enhancing the overall design capabilities of complete-solver LHHs like ours.

\paragraph{KnoBase}
KnoBase is a text-based knowledge repository constructed from task-specific search tags produced by the LLM. LLM researchers use these tags to gather and summarize prior expert knowledge, which is then fed back into the LLM context. This makes the heuristic design process more informed and reduces reliance on the LLM's latent parameter memory.

\section{Experiments}
\label{sec:experiment}

\begin{table*}[h]
\centering
\caption{Detailed Budget Settings for Comparison with LHHs (Evolving Stage).}
\label{tab:budget}
\begin{tabular}{@{}lllllll@{}}
\toprule
\textbf{Section} & \textbf{Problem} & \textbf{LLM Model} & \textbf{Budget Type} & \textbf{Budget} & \textbf{Attempts} & \textbf{Metric} \\ \midrule
Section~\ref{sec:traditionalBench} & TSP & DeepSeek-V3 & Time & 40min & 5 & Best \& Average  \\
Section~\ref{sec:traditionalBench} & BPP & DeepSeek-V3 & Time & 60min & 5 & Best \& Average  \\
Section~\ref{sec:traditionalBench} & CAF & DeepSeek-V3 & Time & 100min & 5 & Best  \\
\midrule
Section~\ref{sec:hybridAlgorithm} & MIS (w/ RLSA) & DeepSeek-V3 & Time & 4h & 3 & Best$^*$  \\ 
Section~\ref{sec:hybridAlgorithm} & MIS (w/ KaHIP \& ARW) & DeepSeek-V3 & Time & 4h & 3 & Best  \\ 
Section~\ref{sec:hybridAlgorithm} & CVRP & DeepSeek-R1 & Time & 5h & 3 & Best  \\
Section~\ref{sec:hybridAlgorithm} & TSP & DeepSeek-V3 & Time & 1h & 3 & Best  \\
Section~\ref{sec:hybridAlgorithm} & BBOB & DeepSeek-V3 & Time & 18min & 3 & Best \\
\midrule
Section~\ref{sec:morecr} & CVRP & DeepSeek-R1 & Token & 55w & 15 & Best \& Average  \\
Section~\ref{sec:morecr} & TSP & DeepSeek-V3 & Token & 6w & 15 & Best \& Average  \\
\bottomrule
\end{tabular}
\par \footnotesize $*$: In Section~\ref{sec:hybridAlgorithm}, since we are comparing LHHs with SOTA solvers, we only report the best results for comparison.
\end{table*}

In our experiments, we fixed the budgets for LHHs during the evolving stage. Generally, we \textbf{strictly adhere to their original hyperparameter configurations}, maintaining identical proportional relationships while only scaling magnitudes to ensure equivalent evaluation budgets. Besides, all experiments use consistent LLM models and temperature settings. However, due to the various difficulties of different problems, the budgets for different problems are different, as shown in Table~\ref{tab:budget}.

Hardware and hyperparameter details are shown in Appendix~\ref{sec:implementationDetails}. For a given problem, we cover the test instances across all sizes with the same algorithm. The BEAM-generated algorithms are provided in Appendix~\ref{sec:generatedCodes}. 

\subsection{On Traditional Single Function Evaluation}
\label{sec:traditionalBench}

In this section, we compare BEAM with other LHHs and expert-designed heuristics using traditional evaluation settings, averaging the performance of their best-generated heuristics over 5 trials. We use the three most-tested \textbf{single function design} tasks: \textbf{1)} Design penalty heuristics in the Guided Local Search framework for the Traveling Salesman Problem (TSP). \textbf{2)} Design priority functions for the Bin Packing Problem (BPP). \textbf{3)} Design Cost-aware Acquisition Functions (CAF) for Bayesian Optimization (BO). For CO, we compare BEAM with ReEvo, EoH and MCTS-AHD. For CAF, we compare our results with more LHHs and expert-designed EI-cool~\cite{EI-cool} (EI~\cite{EI} + EIpu~\cite{EIpu}).

\begin{table*}[t]
    \centering
    \scriptsize
    \caption{Results on Traditional CO Benchmark (TSP \& BPP Single Function Design).}
        \vspace{4pt}
    \label{tab:traditionalmain}
    \resizebox{0.96\textwidth}{!}{
    \begin{tabular}{l c c c c c c c c c c}
        \toprule
        \multicolumn{1}{c}{\textbf{Methods}} & \multicolumn{2}{c}{\textbf{TSP-100}} & \multicolumn{2}{c}{\textbf{TSP-500}} & \multicolumn{2}{c}{\textbf{Weibull5k}} & \multicolumn{2}{c}{\textbf{Weibull10k}} & \multicolumn{2}{c}{\textbf{Weibull100k}} \\
          & MIN$\downarrow$ & AVG$\downarrow$  & MIN$\downarrow$ & AVG$\downarrow$ & MIN$\downarrow$ & AVG$\downarrow$  & MIN$\downarrow$ & AVG$\downarrow$  & MIN$\downarrow$ & AVG$\downarrow$ \\
        \midrule
        ReEvo~\cite{ReEvo} & 0.01\% & 0.03\% & 1.00\% & 1.07\% & 2.83\% &3.36\% & 2.71\% &3.33\% & 2.38\% &3.07\%   \\
        EoH~\cite{EoH} & 8.39e-3\% & 0.01\% & \textbf{0.85\%} & 0.96\%  & 3.13\% & \textbf{3.18\%} & 2.90\% &\textbf{3.02\%}  & 2.75\% & 2.87\%   \\
        MCTS-AHD~\cite{MCTSLLM} & 0.02\% & 0.04\% & 0.99\% & 1.12\%  &4.25\% &4.25\% &4.06\%  &4.06\%  &3.88\% &3.88\%   \\
        \hdashline
        BEAM (ours) & \textbf{2.63e-3\%} & \textbf{6.77e-3\%} & 0.88\% & \textbf{0.95\%}  & \textbf{2.58\%} & 3.26\% & \textbf{1.99\%} & 3.04\% & \textbf{1.82\%} & \textbf{2.86\%}   \\
        \bottomrule
    \end{tabular}
    }
\end{table*}

\begin{table*}[tb!]
\centering
\caption{Results on Traditional CAF Benchmark. $^*$Best heuristic from their repository; MH: Manually-designed Heuristic; TH: Trigonometric-Hump, Styblinski: Styblinski-Tang, HM: Hartmann.}
\label{tab:CAF}
\resizebox{\textwidth}{!}{
\begin{tabular}{lccc cc cc cc cc}
\toprule
\textbf{Methods} & \multicolumn{1}{c}{\textbf{Type}}  & \multicolumn{2}{c}{\textbf{Griewank}$\downarrow$} & \multicolumn{2}{c}{\textbf{Rosenbrock}$\downarrow$} & \multicolumn{2}{c}{\textbf{Levy}$\downarrow$} & \multicolumn{2}{c}{\textbf{TH}$\downarrow$} & \multicolumn{2}{c}{\textbf{Styblinski}$\downarrow$}  \\
 & & C=12 & C=120 & C=12 & C=120 & C=12 & C=120 & C=12 & C=120 & C=12 & C=120  \\
\midrule
EI-cool &MH & 0.78  & 0.18 & 7.14 & \cellcolor{gray!20}1.85 & \cellcolor{gray!20}0.43 & 9.14e-4 & 1.78 & 1.05e-3 &11.88 & 3.92e-3  \\

EvoCAF$^*$ & LHH & \cellcolor{gray!20}1.06 & 0.13 & \textbf{4.75} &\textbf{0.05} & 0.11 & 1.77e-3 & 0.24 &1.67e-3 & 1.93 & \cellcolor{gray!20}0.02  \\
MCTS-AHD$^*$ & LHH & 0.56 & \cellcolor{gray!20}0.22 & \cellcolor{gray!20}19.81 & 0.48 & \textbf{0.06} & \cellcolor{gray!20}2.67e-3 & \textbf{0.08} & \cellcolor{gray!20}3.34e-3 & \textbf{0.69} & 7.10e-3    \\

AlphaEvolve & LHH &0.89 &0.19 &15.93 &0.14 &0.15 &2.18e-3 &\cellcolor{gray!20}1.89 &3.06e-3 &\cellcolor{gray!20}14.82 &0.02 \\
\textbf{BEAM (Ours)} & LHH & \textbf{0.49} & \textbf{0.13} &9.81  &0.41 & 0.26 & \textbf{6.99e-4} & 1.68 & \textbf{2.70e-4} & 1.25 & \textbf{1.54e-3}  \\
\hdashline
\midrule
\textbf{Methods}& \multicolumn{1}{c}{\textbf{Type}} & \multicolumn{2}{c}{\textbf{HM3D}$\downarrow$} & \multicolumn{2}{c}{\textbf{Powell}$\downarrow$} & \multicolumn{2}{c}{\textbf{Shekel}$\downarrow$} & \multicolumn{2}{c}{\textbf{HM6D}$\downarrow$} & \multicolumn{2}{c}{\textbf{Cosine8}$\downarrow$}\\
 & & C=12 & C=120 & C=12 & C=120 & C=12 & C=120 & C=12 & C=120 & C=12 & C=120  \\
\midrule
EI-cool &MH & 1.36e-2 & \textbf{9.27e-5} & \cellcolor{gray!20}144.62 & \cellcolor{gray!20}6.91 & 8.83 & 7.25 & 1.25 & 0.06 & 1.18 & 0.38 \\

EvoCAF$^*$ & LHH &\cellcolor{gray!20}5.93e-2  &2.51e-3 & 70.36 & \textbf{0.04} & \cellcolor{gray!20}9.12 & 0.47 & \cellcolor{gray!20}1.84 & \textbf{0.01} &\cellcolor{gray!20}1.57 & \textbf{0.03} \\
MCTS-AHD$^*$ & LHH & 1.39e-2 &2.06e-3 &\textbf{15.78}  & 0.18 & \textbf{7.46} & \textbf{0.15} & 1.46 &\cellcolor{gray!20}0.45 &0.84 & 0.06\\
AlphaEvolve & LHH & 3.76e-2 &\cellcolor{gray!20}3.45e-3 &82.51 &6.32 &8.90 &\cellcolor{gray!20} 7.82 &1.03 &0.10 &\textbf{0.79} &\cellcolor{gray!20} 0.42 \\
\hdashline

\textbf{BEAM (Ours)} & LHH& \textbf{1.27e-2} & 3.95e-4 & 39.62 & 2.29 & 8.79 & 5.61 & \textbf{1.00} & 0.11 &0.97 & 0.39 \\

\bottomrule
\end{tabular}
}
\end{table*}

\textbf{Main Results.}
As shown in Table~\ref{tab:traditionalmain}, BEAM demonstrates strong overall performance while showing slight tendencies of overfitting in TSP and BPP.
Note that EoH is worse than their published results\cite{EoH} since we control the budget for running EoH. While their paper shows a final heuristic with 0.6\% gap on Weibull5k, we cannot reproduce the result even if we triple the budget (\(>2\%\) gap). On CAF benchmarks (Table~\ref{tab:CAF}), BEAM outperforms AlphaEvolve in most datasets within same evolve budget. Other LHHs in the CAF experiment aren't reimplemented and we simply test their best heuristic in their repository, so the budget is unknown for those LHHs. 
While our framework isn't designed for single-function generation tasks - and consequently may introduce unnecessary complexity for such tasks - it nonetheless delivers competitive results.

\subsection{On Proposed KA-guided Evaluation}
\label{sec:hybridAlgorithm}

We conduct experiments using our KA-guided evaluation pipeline. For CO problems, we implement runtime control by requiring LLM-generated algorithms to include a time-checking mechanism. This is implemented via Python's \texttt{time.time()} function with a timeout parameter passed to the generated code. Runtime budgets for different problem sizes are listed in the table captions. In this section, we report the best results after three trials. The results are shown in Figure~\ref{fig:algorithmshoulian},  Table~\ref{tab:CombinedTable} and Table~\ref{tab:BBOBTable}.

\begin{table*}[t]
    \centering
    \scriptsize
    \caption{Performance comparison on MIS, CVRP, and TSP.}
    \label{tab:CombinedTable}
    \scriptsize
    \begin{tabular}{l l ccc ccc ccc ccc}
        \toprule
        \textbf{Methods} & \textbf{Type} & \multicolumn{3}{c}{\textbf{RB-200-300 (t=5s)}} & \multicolumn{3}{c}{\textbf{RB-800-1200 (t=60s)}} & \multicolumn{3}{c}{\textbf{SATLIB (t=60s)}} \\
        & & T$^*$ & OBJ$\uparrow$ & GAP$\downarrow$ & T & OBJ$\uparrow$ & GAP$\downarrow$ & T & OBJ$\uparrow$ & GAP$\downarrow$  \\
        \midrule
        KaMIS (ReduMIS, 60s) & GA & 15k & \textbf{20.09} & 0.00\% & 15k & 43.00 & 0.00\%  & 15k & \textbf{425.95} & 0.00\%  \\
        \midrule
        RLSA & SA & 9k &19.92  & 0.85\%    & 60k & 39.79 & 7.47\%   & 60k & 411.81 &3.32\% \\
        ReEvo w/ RLSA & LHH & 75 &19.99 &0.50\% & 150 &40.79 &5.03\%   &150 & 423.61 & 0.55\%  \\
        EoH w/ RLSA & LHH & 75 &\textbf{20.05} &0.18\%  &150 &41.21 & 4.04\%  & 150 & 424.20 & 0.41\% \\
        MCTS-AHD w/ RLSA & LHH & 75 &20.01 &0.39\%  &150 &41.13 &4.35\%   & 150 &423.96  &0.47\%  \\
        \hdashline
        BEAM w/ RLSA & LHH & 75 &20.05 &0.19\%  & 150 & \textbf{41.65} & 3.05\%  &150  &\textbf{424.24} & 0.40\% \\
        \midrule
        ARW & LS &500k &\textbf{20.09} &0.00\% &2m &42.68 & 0.75\% & 18m & 425.51 & 0.10\% \\
        KaMIS (EvoMIS) & GA &15k &20.09 &0.01\% &15k &42.97 &0.06\% & 15k & 425.95 & -1.44e-3\% \\
        EoH w/ KaHIP\&ARW & LHH  &15k &\textbf{20.09} &0.00\% & 15k  &43.01  & -0.03\% &15k &\textbf{425.95}  & -1.45e-3\% \\
        MCTS-AHD w/ KaHIP\&ARW & LHH  &15k &\textbf{20.09} &0.00\% & 15k  &42.97  &0.08\%  &15k &425.91  &0.01\%  \\
        AlphaEvolve w/ KaHIP\&ARW & LHH  &15k &20.09 &0.01\% & 15k  &42.92  &0.19\%  &15k &425.69  &0.06\%  \\
        \hdashline
        BEAM w/ KaHIP\&ARW & LHH  &15k &\textbf{20.09} &0.00\% & 15k &\textbf{43.03} &-0.06\% & 15k &425.95  &-1.93e-5\% \\
        \bottomrule
    \end{tabular}
    \par \scriptsize *: We control T (the iterations of local search algorithms) to ensure a similar runtime for fair comparison.

    \scriptsize
    \begin{tabular}{l l ccc ccc ccc ccc}
        \toprule
        \textbf{Methods} & \textbf{Type} & \multicolumn{2}{c}{\textbf{CVRP-100 (t=20s)}} & \multicolumn{2}{c}{\textbf{CVRP-200 (t=60s)}} & \multicolumn{2}{c}{\textbf{CVRP-500 (t=300s)}} \\
        & & OBJ$\downarrow$ & GAP$\downarrow$  & OBJ$\downarrow$ & GAP$\downarrow$ & OBJ$\downarrow$ & GAP$\downarrow$  \\
        \midrule
        HGS~\cite{HGS} & GA & \textbf{15.56} & 0.00\%  & \textbf{19.63} & 0.00\%  & \textbf{37.15} & 0.00\%  \\
        \midrule
        Split~\cite{Split} \& LS$^*$ & LS &15.65  & 0.56\%   & 20.00 & 1.89\%   &38.56  & 3.80\% \\
        ReEvo w/ Split \& LS & LHH &15.62 &0.37\% &19.79 &0.79\% &37.71 &1.52\% \\
        MCTS-AHD w/ Split \& LS & LHH & 15.58 & 0.13\%   &19.74  & 0.56\%   &37.63  & 1.29\% \\
        EoH w/ Split \& LS & LHH & 15.59 & 0.22\%   &19.74  & 0.57\%   &37.55  & 1.09\% \\
        \hdashline
        BEAM w/ Split \& LS & LHH &\textbf{15.57}  & 0.09\%   &\textbf{19.70}  & 0.38\%   & \textbf{37.47} & 0.86\% \\
        \bottomrule
    \end{tabular}
    \par \scriptsize*: We perform the two algorithms on random permutations. The runtime is controlled to match its counterparts.
    \scriptsize
    \begin{tabular}{l l ccc ccc ccc ccc}
        \toprule
        \textbf{Methods} & \textbf{Type} & \multicolumn{2}{c}{\textbf{TSP-50 (t=5s)}} & \multicolumn{2}{c}{\textbf{TSP-100 (t=15s)}} & \multicolumn{2}{c}{\textbf{TSP-500 (t=40s)}} \\
        & & OBJ$\downarrow$ & GAP$\downarrow$  & OBJ$\downarrow$ & GAP$\downarrow$ & OBJ$\downarrow$ & GAP$\downarrow$  \\
        \midrule
        EACO-EDM~\cite{ReEvo} & ACO &5.73  & 0.00\%  &8.13 & 0.00\%  &19.80 & 0.00\%  \\
        \midrule
        EoH~\cite{EoH} w/ EDM~\cite{ReEvo} & LHH &5.76 &0.52\% &8.13 &-3.7e-4\% &18.11 &-8.53\% \\
        \hdashline
        BEAM w/ EDM~\cite{ReEvo} & LHH &\textbf{5.73} &-0.10\% &\textbf{7.90} &-2.83\% &\textbf{17.69} &-10.66\% \\
        \bottomrule
    \end{tabular}
\end{table*}

\begin{table*}[t]
\centering
\caption{Different LHHs on BBOB evaluation. $^*$From LLaMEA's repository~\cite{LLaMEARepo} (we test the ERADS function which is claimed to be its best design).}
\label{tab:BBOBTable}
\resizebox{0.8\linewidth}{!}{
\begin{tabular}{lcccccc}
    \toprule[1.2pt]
    \multirow{2}{*}{\textbf{Methods}} & \textbf{Rastrigin} & \textbf{Rosenbrock} & \textbf{Sphere} & \textbf{Ackley} & \textbf{Griewank} & \textbf{Average} \\
     & GAP$\downarrow$ & GAP$\downarrow$ & GAP$\downarrow$ & GAP$\downarrow$ & GAP$\downarrow$ & GAP$\downarrow$ \\
    \midrule
    LLaMEA~\cite{LLaMEA}$^*$ & 0.995 & \textbf{0.000} & \textbf{0.000} & 4.4e-16 & 0.007 & 0.201 \\
    ReEvo~\cite{ReEvo} & \textbf{0.002} & 4.785 & \textbf{0.000} & 1.1e-5 & \textbf{1e-6} & 0.957 \\
    EoH~\cite{EoH} & 10.519 & 14.804 & 0.543 & 0.714 & 3.5799 & 6.032 \\
    \hdashline
    LlaMEA-HPO & 1.512 & 0.419 & 5.5e-10 & 6.9e-5 & 0.001 & 0.386 \\
    BEAM & 0.026 & \textbf{0.000} & \textbf{0.000} & \textbf{0.000} & 0.007 & \textbf{0.007} \\
    \bottomrule[1.2pt]
\end{tabular}}
\end{table*}

\begin{table}[t]
\centering
\caption{ Ablation study on Adaptive memory, education, and model-generalization (TSP is tested on TSP-500, MIS is tested on RB 800-1200, CVRP is tested on CVRP-500, CAF is tested on Ackley \& Rastrigin.)}
\label{tab:ablation}
\vspace{4pt}
\begin{tabular}{@{}l ccc@{}}
\toprule
\textit{Adaptive Memory} & TSP & CVRP & CAF \\
\midrule
BEAM & \textbf{-9.55\%} & \textbf{0.86\%} & \textbf{3.46\%} \\
BE & -8.12\% & 0.89\% & 4.41\% \\
\toprule
\textit{Education Method} & MIS & CVRP & CAF \\
\midrule
One-Shot & 3.63\% & 1.07\% & \textbf{5.12\%} \\
MCTS & \textbf{3.05\%} & \textbf{0.86\%} & 8.17\% \\
\toprule
\textit{Model Generalization} & TSP-50 & TSP-100 & TSP-500 \\
\midrule
Deepseek-V3 & \textbf{0.00\%} & \textbf{0.00\%} & \textbf{0.00\%} \\
GPT 3.5 turbo & 0.38\% & 2.64\% & 6.60\% \\
GPT 4o mini & 0.19\% & 0.13\% & 0.12\% \\
\bottomrule
\end{tabular}
\end{table}

\textbf{Main Results.} For CO problems, BEAM surpasses KaMIS without reduction operations, achieves results close to HGS, and outperforms existing LHHs, demonstrating its superiority in complex code generation. Note that in TSP, since the ACO in ReEvo's repository is a general framework without 2-opt~\cite{2OPT}, a robust ACO framework integrated with 2-opt can easily surpass EACO-EDM. BEAM and EoH both implement 2-opt. However, EoH fails to consistently outperform EACO-EDM across all datasets, suggesting its suboptimal ACO design. For Continuous Optimization problems, results on BBOB show that BEAM can also achieves near-SOTA performance in continuous domains. Further insights derived from BEAM-designed solvers can be found in Appendix~\ref{sec:codeintro}.

\begin{figure}[tb!]
\centering
\includegraphics[width=0.75\linewidth]{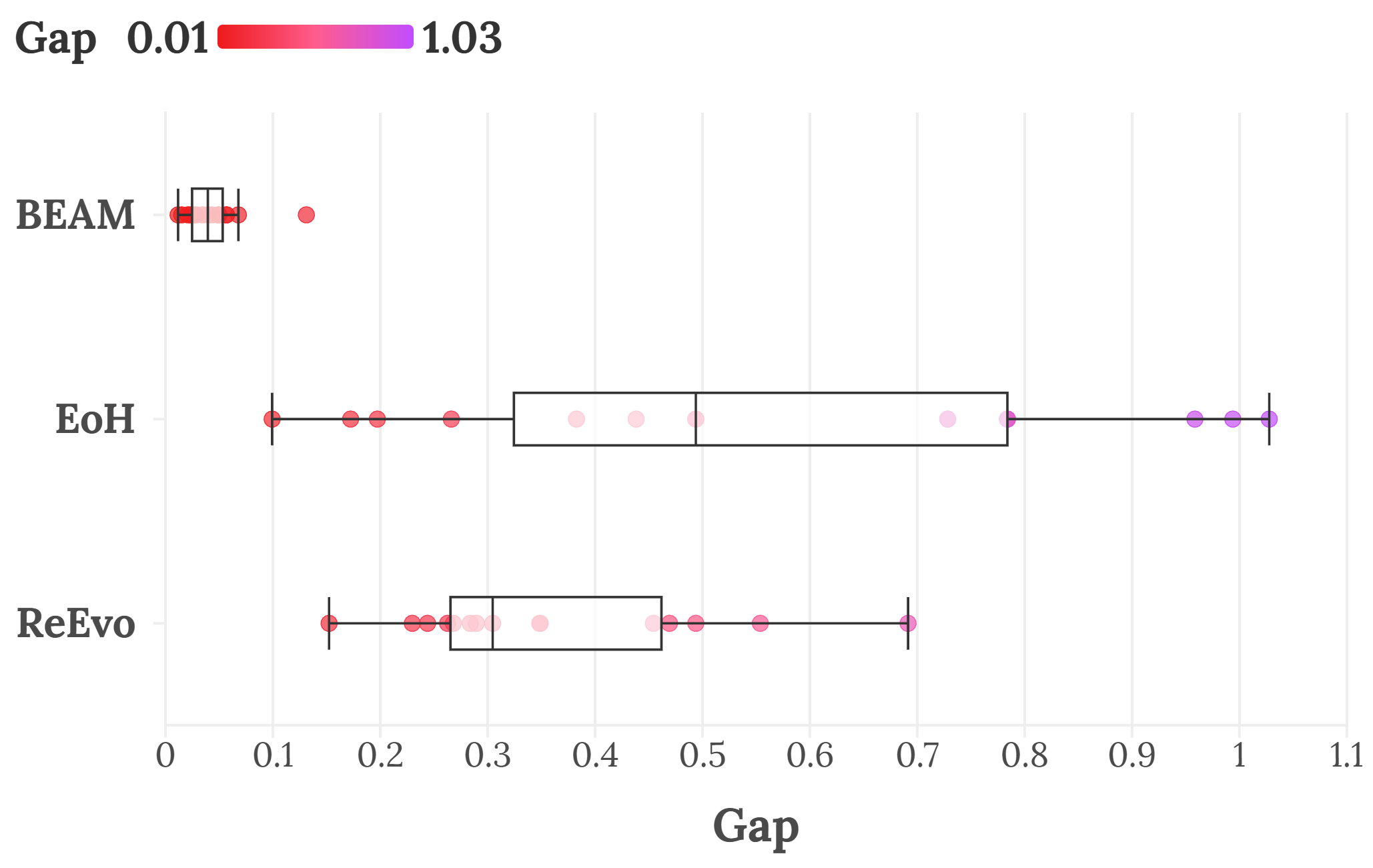}
        \vspace{-10pt}
\caption{Best individual distribution. Label `Gap' is the performance of the code generated by the models. It shows that the code generated by BEAM has the best average performance and the smallest variance among the three models.}
\label{fig:bestIndiDistribution}
\end{figure}

\subsection{More Comparative Results}
\label{sec:morecr}

\paragraph{Best Individual Distribution.} We execute each of the three LHHs 15 times on CVRP (a comparatively complex task) using the same evaluation dataset and record their best fitness values from each run. From Fig.~\ref{fig:bestIndiDistribution}, we can see that the average peak performance of BEAM far exceeds that of ReEvo and EoH. Beside, EoH shows the poorest stability confirming HSEvo's findings~\cite{HSEvo}, and BEAM achieves exceptional stability. 

\begin{figure}[tb!]
    \centering
    \begin{minipage}[t]{0.48\textwidth}
        \centering
        \includegraphics[width=\linewidth]{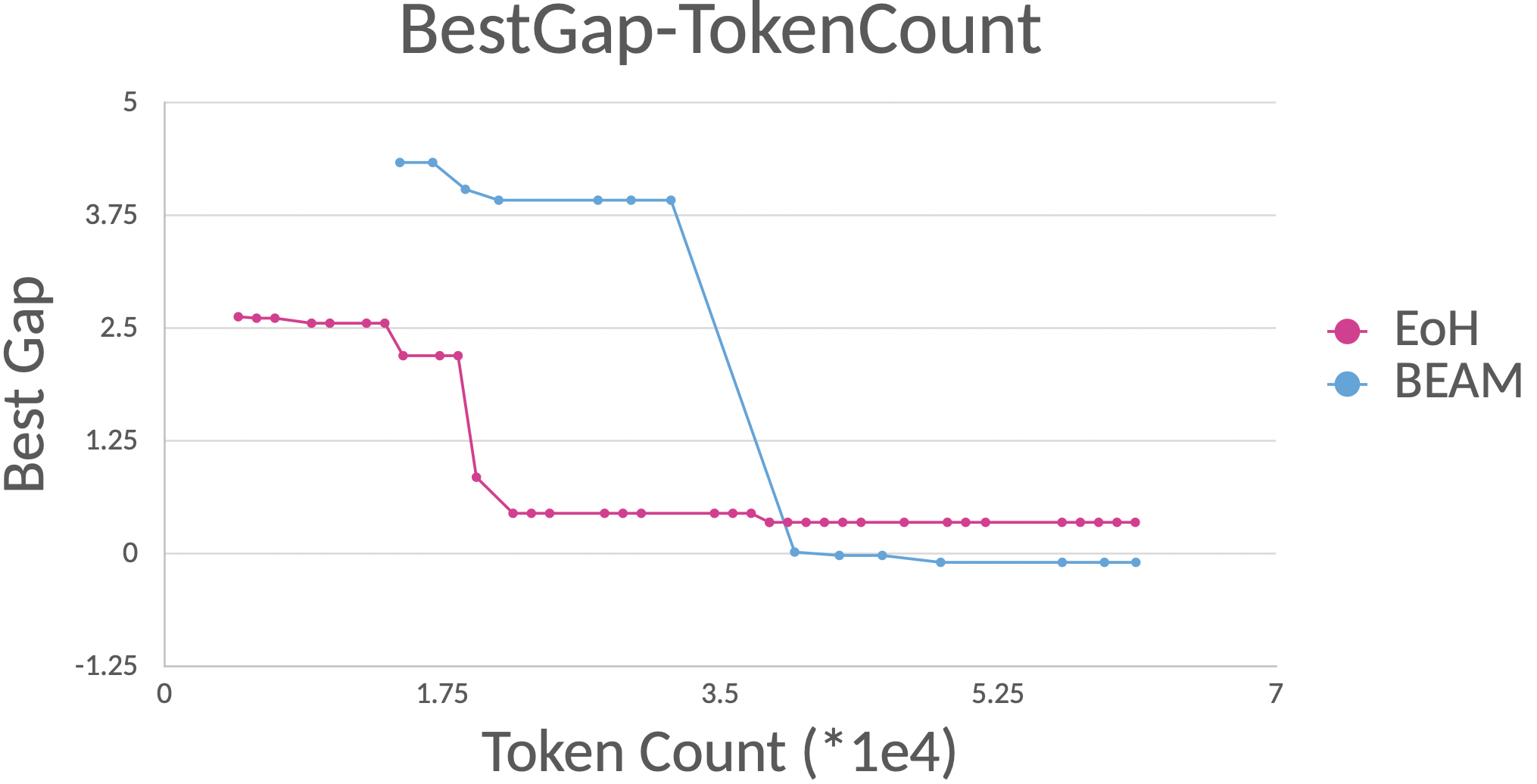}
        \caption{Gap-Token Tendency  (TSP).}
        \label{fig:beatgap-tokencount-tsp}
    \end{minipage}
    \hfill 
    \begin{minipage}[t]{0.48\textwidth}
        \centering
        \includegraphics[width=\linewidth]{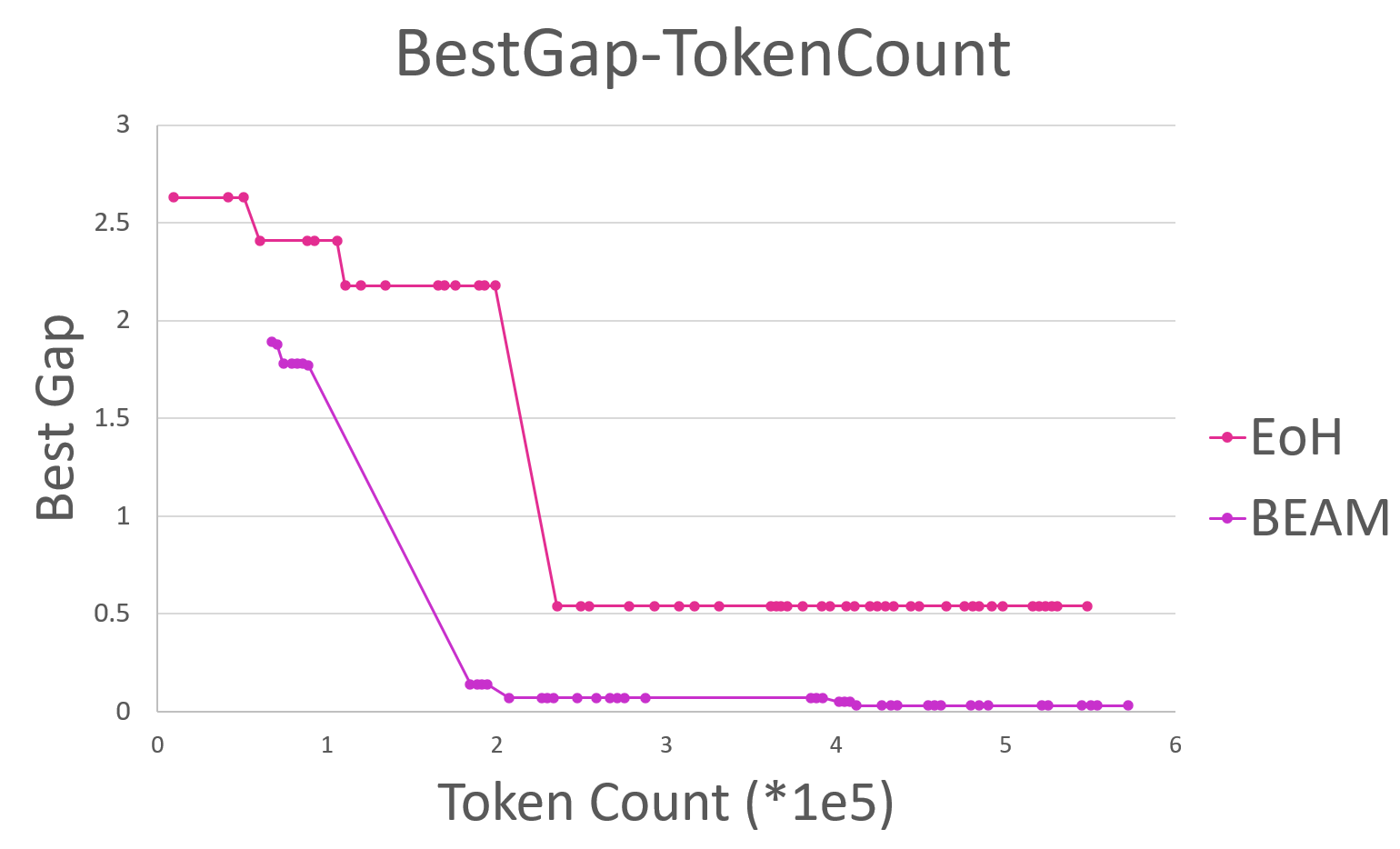}
        \caption{Gap-Token Tendency (CVRP).}
        \label{fig:beatgap-tokencount-cvrp}
    \end{minipage}
\end{figure}

\paragraph{Evolution Curve}
We analyze the median-performing evolutionary processes from 15 runs, tracking how their fitness values scale with token counts. Figures~\ref{fig:beatgap-tokencount-tsp} and~\ref{fig:beatgap-tokencount-cvrp} show the performance gap (y-axis, lower is better) versus cumulative token consumption (x-axis) for TSP and CVRP respectively. Each curve represents one LHH framework's evolutionary trajectory, where points indicate when new individuals are generated and evaluated.
Among them, BEAM has the greatest improving ability.
However, due to its bi-layer structure, it needs the largest number of tokens to generate its first heuristic individual. 
In CVRP (a more complex task), the initial vacancy is because most of the initial codes suffer from execution errors.

\begin{figure}[tb!]
\centering
\includegraphics[width=0.75\linewidth]{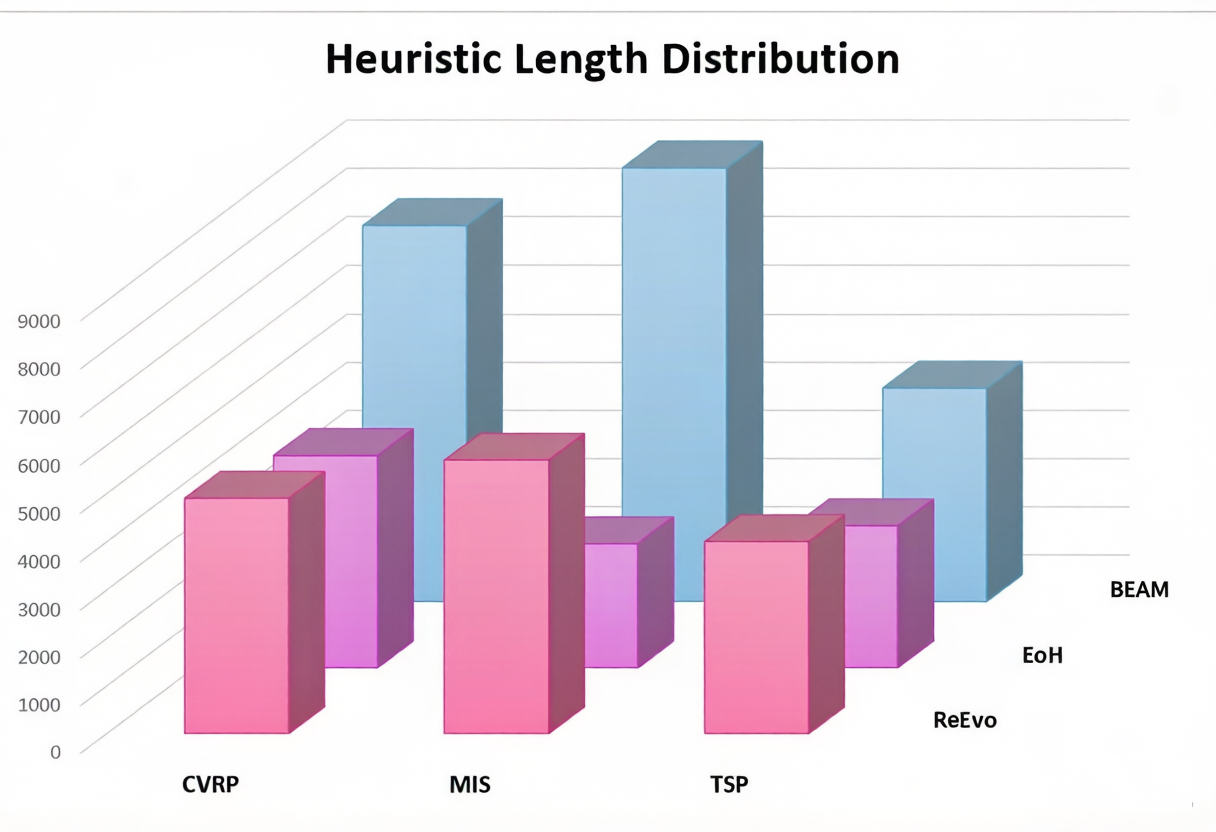}
        \vspace{-10pt}
\caption{Heuristic Length Distribution. BEAM consistently produces the longest and most complex heuristics, which is one reason for its superior performance.}
\label{fig:codeLength}
\end{figure}

\paragraph{Differences of Generated Heuristic}
As illustrated in Fig.~\ref{fig:codeLength}, BEAM consistently produces the longest and most complex heuristics (same requirement prompts). The heuristics are also more robust compared to other LHHs, with detailed code comparison provided in Appendix~\ref{sec:generatedcompare}.

\subsection{Ablation Study}

\begin{figure}[h]
    \centering
    \includegraphics[width=0.8\linewidth]{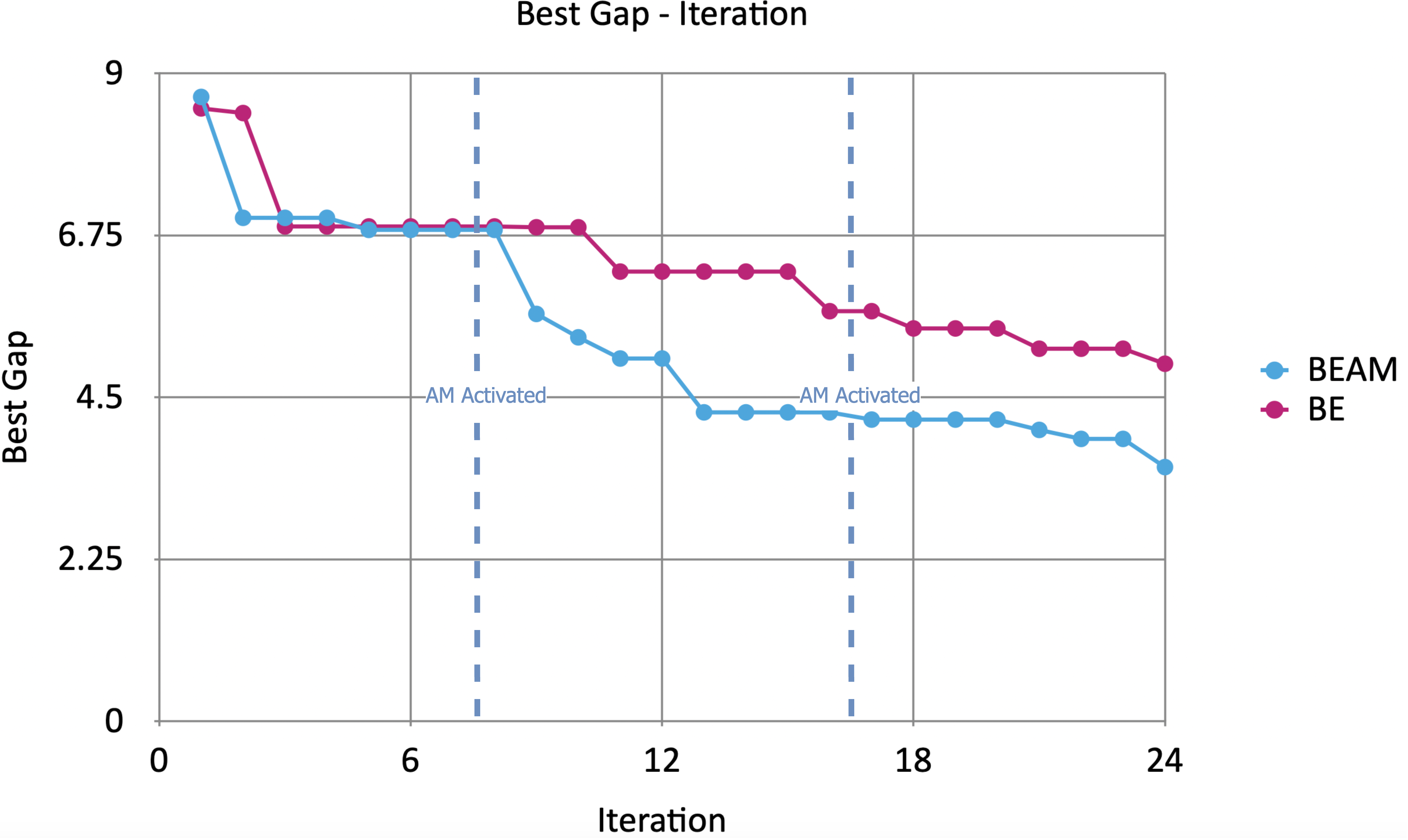}
    \caption{BEAM vs. BE: Best Gap - Iteration.}
    \label{fig:beamvsbe}
\end{figure}

\paragraph{On Adaptive Memory}
Table~\ref{tab:ablation} shows that our proposed Adaptive Memory is effective. It also enhances evolution stability. Fig.~\ref{fig:beamvsbe} compares the iteration curves of BEAM and BE (BEAM without AM) during the evolution process. We observe that BEAM not only achieves higher peak performance but also exhibits better stability. To quantify it, Table~\ref{tab:stability} reports the mean and variance of the gaps at the final iteration over 5 independent runs. The results show that BEAM achieves both a higher mean and a lower variance, implying more robust and reliable convergence.

\begin{figure}[h]
    \centering
    \begin{minipage}{0.38\textwidth}
        \centering
        \begin{tabular}{lcc}
        \toprule
        \textbf{Methods} & \textbf{AVG}$\downarrow$ & \textbf{VAR}$\downarrow$ \\
        \midrule
        BEAM & \textbf{3.46} & \textbf{0.01} \\
        BE & 4.41 & 0.19 \\
        \bottomrule
        \end{tabular}
        \captionof{table}{BEAM vs. BE: Average and Variance of Gaps.}
        \label{tab:stability}
    \end{minipage}
\end{figure}

\paragraph{On Individual Education Method} We test on two individual education methods and the results are shown in Table~\ref{tab:ablation}. We disable calibration for fairness. MCTS outperforms One-Shot except in CAF, where the objective is easy and the importance of structure outweighs functions.

\paragraph{On KA} The LLM-retrieved part of HeuBase is proved effective in BBOB, with CMA-ES being installed and utilized (see~\ref{sec:generatedCodes}). For KnoBase tests, we selected a Parallel Machine Scheduling Problem (PMSP) variant reformulated by ZeroBubble~\cite{ZeroBubble} used by DualPipe~\cite{V3Report} in training DeepSeek-V3. Different from traditional 1F1B~\cite{PipeDream}, this reformulated problem is fairly difficult with loads of constraints (see~\ref{sec:formalProblem}). With KnoBase, BEAM incorporated the Sweep Line Algorithm and generated more constraint-satisfying initial solutions.

\paragraph{Model Generalization}

To evaluate the dependency on LLM size, we test smaller models on TSP, with results shown in Table~\ref{tab:ablation}. The result shows that despite slight increase in gaps, our BEAM still generates high-quality code with small models, showing that the demand and performance of LLM is not the decisive part of our BEAM.

\section{Conclusion and Future Work}
\label{sec:conclusion}

In this paper, we propose BEAM, a Bi-layer structure that separates the heuristic design into two layers: the exterior layer for high-level algorithmic design and the interior layer for detailed function implementation. This structure, integrated with MCTS-based function selection and Adaptive Memory, enables the generation of high-quality, complex heuristics for both continuous and combinatorial optimization problems. We also introduce a Unified KA pipeline for more meaningful LHH evaluation. Experiments show that BEAM outperforms existing LHHs and SOTA solvers. Future work includes expanding BEAM to more complex domains and exploring more efficient ways for KA.

\bibliographystyle{IEEEtran}
\bibliography{ref}

\appendices
\appendix

\section{Extended Discussions}
\label{sec:extendedDiscussions}

\subsection{Additional Comparison}
\label{sec:detailedComparison}

Our usage of MCTS is fundamentally different from PoH and MCTS-AHD as detailed in Table~\ref{tab:RLComparison}.

\subsection{Diversity Discussions}
\label{sec:diversityDiscussions}
Counterintuitively, the introduction of HeuBase enhances rather than diminishes the diversity of LLM-generated functions. This phenomenon can be attributed to two primary reasons: \textbf{1)} \textit{diverse application} of components (e.g. using KaHIP for either initialization or crossover), and \textbf{2)} \textit{stochastic selection} Table~\ref{tab:sef} shows the selection frequency and drop rate in an extra MIS experiment with 50 samples, showcasing diversity.

\begin{table}[h]
\centering
\caption{Selection Frequency of HeuBase Functions.}
\label{tab:sef}
\begin{tabular}{lc}
\toprule
Function & Selection Frequency \\
\midrule
\texttt{kaffpa} & 82\%  \\
\texttt{node\_separator} & 56\%  \\
\texttt{arw} & 80\%   \\
\texttt{arw\_1iter} & 86\%   \\
\texttt{deterministic\_rounding} & 14\%   \\ 
\bottomrule
\end{tabular}
\end{table}

\begin{table*}[ht]
\centering
\caption{Further Comparison.}
\label{tab:RLComparison}
\begin{tabular}{|l|p{8.5cm}|}
\hline
\textbf{Method} & \textbf{MCTS Usage} \\
\hline
MCTS-AHD &   This work just substitutes EoH's evolutionary operator with an MCTS-like strategy for algorithm search, still being a single-layer framework.\\
\hline
PoH  &  The only difference between PoH and MCTS-AHD is that PoH introduced a reflection step similar to ReEvo in the MCTS process.\\
\hline
BEAM (Ours)  &In our work, MCTS is used as the education strategy for an algorithm structure in the interior layer of our bi-layer structure, searching different realization for the unrealized functions of the structure.  \\
\hline
\end{tabular}
\end{table*}

\section{Additional Evaluation Settings}
\label{sec:evaluationSetting}

\subsection{Evaluation Overview}
Table~\ref{tab:bench} documents all evaluation configurations. Note that the first three are chosen from some most-used LHH experiments. We exclude experiments like designing heuristic guide functions for ACO solving problems like TSP and CVRP~\cite{ReEvo, MCTSLLM} since empirical results show that there's no obvious performance difference between LHHs. We also replace the TSP w/ EDM task (described in Section~\ref{sec:hybridAlgorithm}) with the complete TSP solver design since EDM is a comparatively trivial function and LHHs can also design good TSP solver without it.

\begin{table*}[t]
    \centering
    \scriptsize
    \caption{Unified LHH Evaluation Settings with KA.}
    \label{tab:bench}
    \begin{tabular}{lp{1cm}p{0.8cm}p{1.8cm}p{2.2cm}p{1.2cm}p{3cm}}
        \toprule
        \textbf{Prob.} & \textbf{MC} &\textbf{Type} & \textbf{Design Type} & \textbf{Heuristic Type} & \textbf{Dataset Size}  & \textbf{Allowed KA}  \\
\midrule
TSP$^*$  & Easy &GCO & Single Function & Not Restricted & 3*128  & / \\
\midrule
BPP & Easy &CO & Single Function & Not Restricted & 3*5  & / \\
\midrule
CAF & Easy &BO & Single Function & Not Restricted & 10*2*5  & /  \\
\midrule
MIS & Medium &GCO & Hybrid Algorithm & GA & 3*500 & HeuBase (RLSA) \\
\midrule
MIS & Medium &GCO & Hybrid Algorithm & GA & 3*500 & HeuBase (KaHIP, ARW) \\
\midrule
TSP  & Medium &GCO & Entire Algorithm & ACO & 3*128  & / \\
\midrule
BBOB & Medium &BBO & Entire Algorithm & Not Restricted & 5 & HeuBase (LLM-Retrieved), KnoBase \\
\midrule
CVRP & Hard &GCO & Hybrid Algorithm & Not Restricted & 3*100 & HeuBase (Split, LS) \\
\midrule
PMSP & Hard &CO & Entire Algorithm & Not Restricted & 4*3 & HeuBase (LLM-Retrieved), KnoBase \\
        \bottomrule
    \end{tabular}
    \par \footnotesize *: The target is to design a penalty heuristic within the Guided Local Search framework with \texttt{perturbation\_moves} set to 30 and \texttt{iter\_limit} set to 1200; GCO: Graph Combinatorial Optimization, BO: Bayesian Optimization, BBO: Black Box Optimization, ACO: Ant Colony Optimization; MC: Model Complexity
\end{table*}

\subsection{Formal Problem Descriptions}
\label{sec:formalProblem}
\paragraph{Traveling Salesman Problem (TSP).} 
Given a complete graph $G=(\mathcal{V},\mathcal{E})$ with $|\mathcal{V}|=N$ nodes and a symmetric cost matrix $\mathbf{C} \in \mathbb{R}^{N \times N}$ where $\mathbf{C}_{ij} = \mathbf{C}_{ji}$ denotes the cost of traveling between nodes $i$ and $j$, the objective is to find a Hamiltonian cycle $\tau = (i_1, i_2, \dots, i_N, i_1)$ that starts and ends at the same node, visits all other nodes exactly once, and minimizes the total tour cost: $\sum_{k=1}^{N-1} \mathbf{C}_{i_k i_{k+1}} + \mathbf{C}_{i_N i_1}$, with $\mathbf{C}_{ii} = 0$ for all $i \in \mathcal{V}$.

\paragraph{Online Bin Packing Problem (BPP)} 
Given a sequence of items $\{x_1, x_2, \dots, x_N\}$ with sizes $x_i \in (0,1]$ arriving one by one, the goal is to assign each item to a bin upon arrival without knowledge of future items. Each bin has a capacity of 1, and no bin may exceed this capacity. The objective is to minimize the total number of bins used to pack all items.

\paragraph{Cost-aware Acquisition Functions (CAF) for Bayesian Optimization (BO)} In Bayesian Optimization (BO), we aim to optimize an unknown function $ f: \mathcal{X} \to \mathbb{R} $ with evaluation cost $ c(x) > 0 $ varying over $ x \in \mathcal{X} $. A CAF is defined as $\alpha_{\text{CAF}}(x) := \frac{\alpha(x)}{c(x)}$ where: $\alpha(x)$ is a standard acquisition function such as Expected Improvement (EI), Upper Confidence Bound (UCB), or Probability of Improvement (PI); $c(x)$ is the cost of evaluating $f$ at point $x$. The next query point is then selected by solving:
$x_{t+1} = \arg\max_{x \in \mathcal{X}} \alpha_{\text{CAF}}(x) = \arg\max_{x \in \mathcal{X}} \frac{\alpha(x)}{c(x)}$
which prioritizes locations that offer the highest expected gain per unit cost.

\paragraph{Maximum Independent Set (MIS)} Given a unweighted graph $G=(\mathcal{V},\mathcal{E})$, an \textit{independent set} $S \subseteq \mathcal{V}$ is a subset of nodes such that no two nodes in $S$ are adjacent. The goal is to maximize $|S|$ s.t. $\forall i, j \in S$, $(i,j) \notin \mathcal{E}$.

\paragraph{Capacitated Vehicle Routing Problem (CVRP)} Given a graph $G=(\mathcal{V},\mathcal{E})$, a depot node $v_0 \in \mathcal{V}$, a cost matrix $\mathbf{C} \in \mathbb{R}^{N \times N}$, a demand vector $\mathbf{d} \in \mathbb{R}_{+}^N$, and a vehicle capacity $Q > 0$, the goal is to plan a set of routes $\mathcal{R}$, each route $r \in \mathcal{R}$ starting and ending at the depot $v_0$, such that each customer node is visited exactly once and the total demand on each route does not exceed $Q$, i.e., $\sum_{i \in r} \mathbf{d}_i \leq Q$. The objective is to minimize the total cost of all routes: $\min\limits_{\mathcal{R}} \sum\limits_{r \in \mathcal{R}} \sum\limits_{(i,j) \in r} \mathbf{C}_{ij}
$.

\paragraph{Black Box Optimization Benchmark (BBOB)} Black‑Box Optimization Benchmarking (BBOB) is COCO’s standard suite of 24 noiseless, single‑objective test functions—provided in dimensions 2, 3, 5, 10, 20, and 40 with multiple randomized instances—to objectively compare black‑box optimizers under a fixed function‑evaluation budget. Performance is measured by the number of evaluations required to reach target accuracies, convergence curves, and success rates across functions of varying separability, conditioning, and multimodality.

\paragraph{Parallel Machine Scheduling Problem (PMSP)} \textit{This problem is reformulated by ZeroBubble~\cite{ZeroBubble}}: Any pass in a pipeline can be uniquely identified by a triple $(i, j, c)$, where $i \in \{1, 2, \dots, p\}$ denotes the stage, $j \in \{1, 2, \dots, m\}$ denotes the microbatch index, and $c \in \{F, B, W\}$ represents the computation type (forward, backward, weight update). $T_{(i,j,c)}$ is the execution time of pass $(i, j, c)$, and $E_{(i,j,c)}$ is its ending time. $\Delta M_{(i,j,c)}$ denotes the memory change incurred by pass $(i, j, c)$. For example, $\Delta M_{(\cdot, \cdot, F)} = M_B$ indicates that the forward pass increases memory usage by $M_B$. Similarly, the backward pass frees $M_B$ while requiring memory for weights $M_W$, hence $\Delta M_{(\cdot, \cdot, B)} = M_W - M_B$, and the weight update consumes $\Delta M_{(\cdot, \cdot, W)} = -M_W$. A binary indicator $O_{(i,j,c) \to (i',j',c')}$ equals 1 if pass $(i,j,c)$ is scheduled before pass $(i',j',c')$, and 0 otherwise. The PMSP is then formulated as a Mixed Integer Linear Programming (MILP) problem:
\begin{equation}
    \begin{aligned}
& \min_{O, E} \quad && \max_{i} \; E_{(i,m,W)} - E_{(i,1,F)} + T_{(i,1,F)} \\
& \text{s.t.} \quad && E_{(i,j,F)} \geq E_{(i-1,j,F)} + T_{\text{comm}} + T_{(i,j,F)} \\
& && E_{(i,j,B)} \geq E_{(i+1,j,B)} + T_{\text{comm}} + T_{(i,j,B)} \\
& && E_{(i,j,c)} \geq E_{(i,j,c')} + T_{(i,j,c)} - O_{(i,j,c) \to (i,j,c')} \cdot \infty \\
& && M_{\text{limit}} \geq \Delta M_{(i,j',c')} + \sum_{j,c} \Delta M_{(i,j,c)} O_{(i,j,c) \to (i,j',c')}
\end{aligned}
\end{equation}

\begin{table}[h]
\centering
\caption{Evaluation Dataset for BEAM.}
\label{tab:eval}
\vspace{4pt}
\begin{tabular}{@{}lll@{}}
\toprule
\textbf{Problem} & \textbf{Dataset} & \textbf{Instances} \\ \midrule
TSP (GLS) &TSP-200 &10 \\
BPP &Weibull5k &10 \\
CAF &Ackley \& Rastrigin &5*2 \\
MIS (w/ RLSA) &RB 200-300 &25 \\
MIS (w/ KaHIP \& ARW) &RB 800-1200 &10 \\
CVRP &CVRP-100 &20 \\
TSP (w/ EDM) &TSP-50 &30 \\
BBOB &Ellipsoidal \& Levy &2 \\ 
PMSP &Randomly-Generated & 5 \\ \bottomrule
\end{tabular}
\end{table}

\subsection{Dataset Details}
Note that here only provides the test dataset. The evaluation dataset during the LHH process isn't restricted for this setting. The evaluation dataset we use is presented in Table~\ref{tab:eval}.

\paragraph{TSP} We follow DIFUSCO~\cite{DIFUSCO} to conduct experiments on \textit{TSP-50}, \textit{TSP-100} and \textit{TSP-500}.
\paragraph{BPP} Following FunSearch~\cite{funsearch}, we used instances sampled from Weibull distributions. Specifically, we generated Weibull 5k, Weibull 10k, Weibull 100k.
\paragraph{CAF} Following MCTS-AHD~\cite{MCTSLLM}, all the dataset are synthetic instances with different landscapes and input dimensions, and we used Ackley and Rastrigin as the evaluation dataset during evolution, so these two aren't included in the table below. We tested with sampling budgets of 12 and 120 to assess the generalizability of all algorithms. All tests take 5 trials.
\paragraph{MIS} We use the Revised Model B (RB) graphs and SATLIB graphs, following~\cite{RLSA,diffuco}. For RB graphs, we use \textit{RB 200-300} for small-scale and \textit{RB 800-1200} for large-scale.
\paragraph{CVRP} We constructed CVRP-100, CVRP-200 and CVRP-500. Following COExpander~\cite{COExpander} and GOAL~\cite{GOAL}, where the coordinates of the depot and clients were sampled from a uniform distribution over the unit square, consistent with the TSP setting.
\paragraph{BBOB} We choose five functions in BBOB: Rastrigin, Rosenbrock, Sphere, Ackley and Griewank, and use the provided extrema points to evaluate.
\paragraph{PMSP} All the data used for evaluation are real-world data directly taken from the ZeroBubble literature~\cite{ZeroBubble}.

\section{Implementation Details}
\label{sec:implementationDetails}

\subsection{Hyperparameter Setting}
\label{sec:hyperset}
Common hyperparameter setting is presented in Table~\ref{tab:comhy}.

In the ablation study on education methods, we bypass KA and calibration, and Table~\ref{tab:ablhy} shows other settings. In the ablation study on AM, we refresh the population for BE (BEAM without AM) by keeping 2 elite algorithms and injecting 13 newly sampled ones at the same intervals as BEAM. In the ablation study on KA, all settings are the same except whether to include KA. In this section, TSP is tested on TSP-500, CVRP is tested on CVRP-500, MIS is tested on RB-800-1200, and CAF is tested on Ackley and Rastrigin (average over the two).

\begin{table}[h]
\centering
\caption{Common Hyperparameter Setting for BEAM.}
\label{tab:comhy}
\begin{tabular}{@{}ll@{}}
\toprule
\textbf{Parameter} & \textbf{Value} \\ \midrule
\texttt{llm\_temperature} & 0.7 (for Fixing) / 1.0 (others) \\
\texttt{crossover\_rate} & 0.7 \\
\texttt{mutation\_rate} & 0.3 \\ 
\texttt{mc\_func\_pop} & 3 \\
\texttt{max\_func\_num} & 4 \\ \bottomrule
\end{tabular}
\par \footnotesize \texttt{mc\_func\_pop}: The number of functions generated during each func\_i generation.
\end{table}

\begin{table}[h]
\centering
\caption{Hyperparamter Setting for Education Method Ablation Study.}
\label{tab:ablhy}
\begin{tabular}{@{}llllll@{}}
\toprule
&\textbf{Methods} & \texttt{iter} & \texttt{ips} & \texttt{mps} &\texttt{mft}  \\ \midrule
&One-Shot &8 &20 &5 &3  \\
&MCTS &3 &5 &3 &3 \\ \bottomrule

\end{tabular}
\par \footnotesize \texttt{ips}: \texttt{init\_pop\_size}, \texttt{mps}: \texttt{max\_pop\_size},\texttt{mft} : \texttt{max\_fix\_try}
\end{table}

\subsection{Hardware Details}

All experimental evaluations, including both the evolutionary optimization process and final performance assessments, are conducted on an Apple M3 CPU. However, algorithms incorporating RLSA are executed on an NVIDIA® GeForce RTX™ 4070 Ti SUPER GPU due to their PyTorch-based computational requirements and algorithms with KaHIP are run on an Intel(R) Xeon(R) Platinum 8558 96-Core Processor CPU since KaHIP doesn't support ARM64 architecture.

\subsection{More Details on MCTS}
\label{sec:mceg}
\begin{figure*}
	\centering
	\includegraphics[width=1\linewidth]{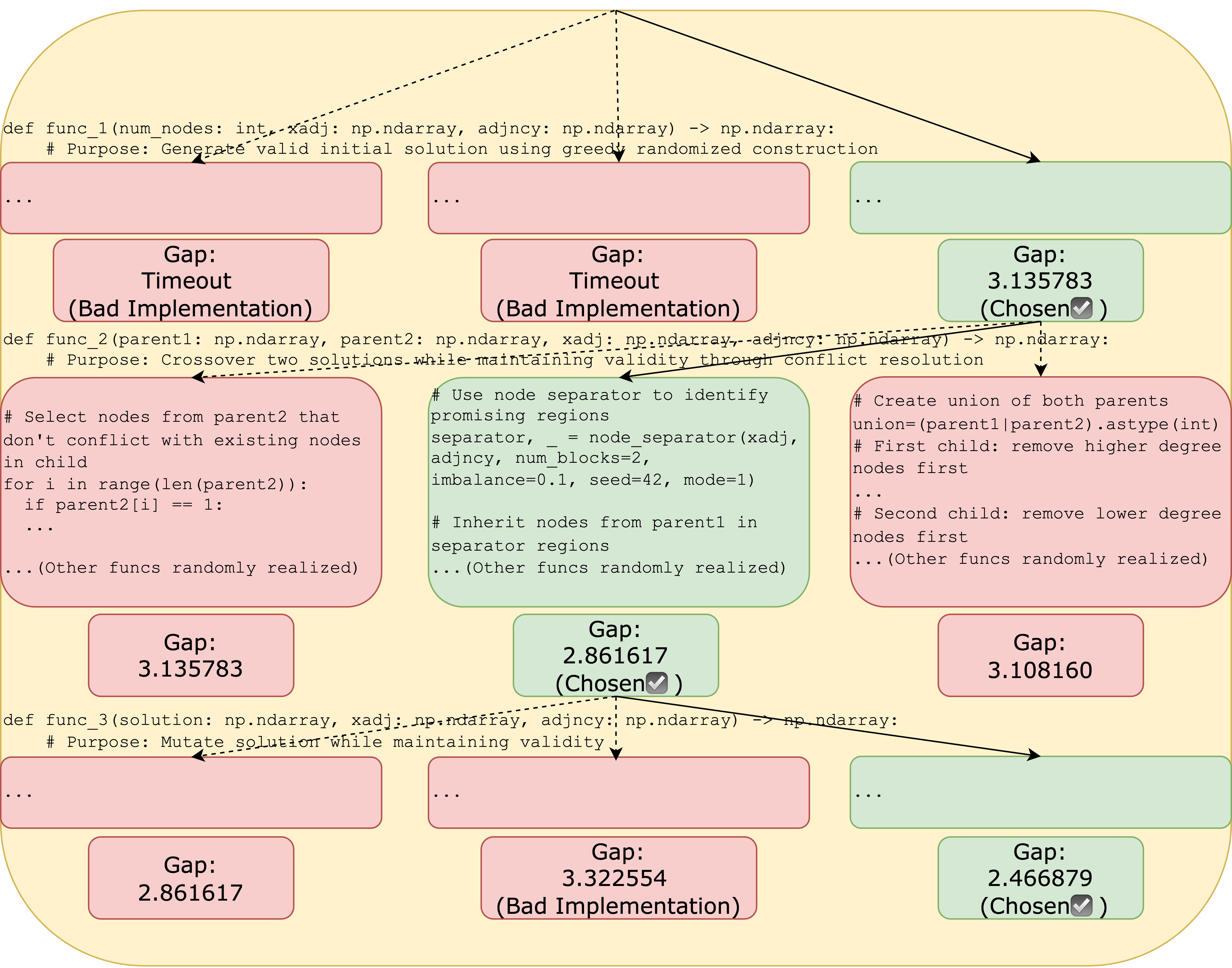}
	\caption{MCTS Example.}
	\label{fig:mceg}
\end{figure*}

Fig.~\ref{fig:mceg} further illustrates the MCTS process. Note that in practice, when prompting the LLM to generate multiple variants for a certain function, we provide its previous designs and tell LLM to ``improve it'' or ``think about a different way to implement this'' to encourage diversity (See Appendix~\ref{sec:usedPrompts} for details). Across multiple runs and different problems, MCTS granted an average performance gain of 46.6\% to each individual.

\section{Used Prompts}
\label{sec:usedPrompts}
\subsection{Common Prompts}
\paragraph{am.txt}
This prompt is used to require the LLM to provide a name and a description for the given function, which will be later added to the Adaptive Memory.

\begin{quote}
\{ function\_code\}

Please read the function code above carefully, and understand what this function achieves. Your job is to give a name of this function and then provide a short description for this function. Note that the description should include the overall description, the meaning of the arguments and the meaning of the output. Your output should follow this:

\texttt{\`}\texttt{\`}\texttt{\`}python

def [the function name] (...copy the provided arguments):

"""

[the overall description]

Args:

    [arg name]: [the meaning of the arguments]

Out:

    [output argument name]: [the meaning of the output]
    
"""
\texttt{\`}\texttt{\`}\texttt{\`}
\end{quote}

\paragraph{ask\_pms\_interval.txt}
This prompt is aimed to require LLM to provide the range of hyperparameters for Calibration part.
\begin{quote}
A heuristic will be given below, all the hyperparameters will be on the top.

You should read the entire code carefully and decide on an interval for each of the hyperparameter.
You should output a python dictionary only, and the dictionary maps from a string (name of the hyperparameter) to a tuple (start, end).

The dictionary must be named pms\_dict.

Format your dictionary as a Python code string: "\texttt{\`}\texttt{\`}\texttt{\`python ... \texttt{\`}\texttt{\`}\texttt{\`}}"
\end{quote}

\paragraph{ask\_pms\_system.txt}
This prompt is the system prompt of ask\_pms\_interval.
\begin{quote}
You are an expert in hyperparameter optimization and you give proper test range for each hyperparameter.

Your suggestions on test range should be in the form of python dictionary.

Your response outputs Python code and nothing else. Format your code as a Python code string: "\texttt{\`}\texttt{\`}\texttt{\`python ... \texttt{\`}\texttt{\`}\texttt{\`}}".
\end{quote}

\paragraph{crossover.txt}
This prompt is the crossover prompt. the \{\} part will be fill in with the parent-structures.
\begin{quote}
\{exterior\_user\_generator\}\\

[Worse code]

\{worse\_code\}\\

[Better code]

\{better\_code\}\\

[Improved code]

Please reflect on why the latter one perform better and write an improved structure according to your reflection. Enclose your code with a Python fenced code block.
\end{quote}

\paragraph{exterior\_user\_generator.txt}
This prompt requires LLM to design the overall structure of the problem. Some requirements and restrictions are provided. Note that we call both AM and HeuBase are called 'Heuristic Database' in the prompt to make a clearer understanding.

\begin{quote}
Now you have to design a novel \{alg\_type\} solving \{problem\}.
\{problem\_description\}

You need to design the overall structure of your \{alg\_type\} as well as thinking detailedly about what you will do in each step and make sure each goal is achievable. Then you must write a piece of code, presenting your algorithm. Here are the requirements of your python code:

1. Put your main structure in the following function:

   \{baseline\}

2. You should look at the *heuristic database* first, and then think:

   - How can you deconstruct the big problem into small subproblems step by step?
   
   - What are the main goal of the subproblems that are needed to complete the heuristic designing?
   Your thoughts should guide you to complete the heuristic design in step 3. You should not think too much of how to realize the subproblems.

3. In this code you needn't implement everything and utilize the idea of **modularization programming**. You can call external functions to represent or compose every subproblems. There are two cases of this external function: (The first case has a higher priority)

   I. The function already exists in the *heuristic database* (which I'll give you later). You can suppose I've already implemented it and directly call it. You must make sure that the function 100\% satisfies your need here. You're encouraged to integrate the existing heuristics into your structure.
   
   **Note**:
   
   - I'll import these heuristics when I test your code, so don't define these heuristics yourself!

   II. The function doesn't exist in the *heuristic database*. Remember to name these external functions 'func\_id' where id is a **number**. It should format as this:
   
   \texttt{\`}\texttt{\`}\texttt{\`}python
   
   def func\_\{\{id\}\}(...) -> ... :
   
     \# Purpose:
     
     pass
     
   \texttt{\`}\texttt{\`}\texttt{\`} 
   
   **Note**: 
   
   - You MUSTN'T implement these functions since I'll let others implement it.
   
   - The purpose should be very clear.
   
   - Don't bother calling these functions if the purpose is too easy!
   
   - There should be at least 1 func\{\{id\}\}, at most {max\_func\_pop} func\{\{id\}\}. The \{\{id\}\} must start from 1.

4. Put the definitions of all the hyperparameters on top of everything. Enclose your hyperparameter list with two "\#Hyperparameter\#". Your hyperparameters must be float or int. If a hyperparameter is int, add "\# int" right after the definition inline.

5. We only have \{timeout\} seconds to perform the algorithm, so you must set your code a {timeout}-second-clock. Include MAX\_TIME = \{timeout\} in your hyperparameter list. In the code, please make frequent check whether the time is up.

6. Let your code print out the best objectives after each iteration.

\{prior\_knowledge\}
\end{quote}

\paragraph{fill\_1func.txt}
This prompt is used in MCTS, where LLM need to temperately fill in only one function to choose the best one to really fill into the structure.
\begin{quote}
An algorithm solving \{problem\} will be given below, with some of the functions realized and others unrealized.

\{problem\_description\}

You are required to complete func\_\{id\}. You should follow the instructions given.
You should output only python code as required.

Your generated code MUST be different from the following code and you should either improve it or think out of box and explore a different way:

\{code\_before\}

**Critical Reminder:**

- You MUST keep ALL existing code, comments, and hyperparameters EXACTLY as provided

- Your response MUST contain the ENTIRE algorithm code

- Only modify the specified func\_\{id\} implementations

- Preserve ALL other code exactly as provided

- Format output as: \texttt{\`}\texttt{\`}\texttt{\`}python [COMPLETE CODE] \texttt{\`}\texttt{\`}\texttt{\`}

{prior\_knowledge}
\end{quote}

\paragraph{fill\_allFunc.txt}
This prompt is used in One-Shot method and MCTS (used after fill\_1func to help decide which is the best function to actually fill into the structure).

\begin{quote}
An algorithm solving **\{problem\}** will be given below, with some of the functions realized and others unrealized.

\{problem\_description\}

You have to implement all the functions named func\_i() where i is a number.

You should follow the instructions given in the code structure when you implement the functions and make sure your code serves the original purpose.

You are encouraged to directly call the functions in the *heuristic database* that will be given below to serve your purpose.

**Critical Reminder:**

- You MUST keep ALL existing code, comments, and hyperparameters EXACTLY as provided

- Your response MUST contain the ENTIRE algorithm code

- Only modify the specified func\_i() implementations

- Preserve ALL other code exactly as provided

- Format output as: \texttt{\`}\texttt{\`}\texttt{\`}python [COMPLETE CODE] \texttt{\`}\texttt{\`}\texttt{\`}

\{prior\_knowledge\}
\end{quote}

\paragraph{fix.txt}
This prompt is used to require LLM to fix the heuristic that reported error.
\begin{quote}
The following code has the following error: \{error\_msg\}. Please fix it.

**Important**: 

- You must **output the entire code** and you must **only** fix the error in the traceback message. Don't fix anything besides the error presented by the error message.

- You can't change MAX\_TIME.

- Don't remove hyperparameters!
\end{quote}

\paragraph{fix\_system.txt}
This prompt is the system prompt of fix.
\begin{quote}
You are an expert in debugging and you output the entire code after debugging.

Your response outputs Python code and nothing else. Format your code as a Python code string: "\texttt{\`}\texttt{\`}\texttt{\`}python ... \texttt{\`}\texttt{\`}\texttt{\`}".
\end{quote}

\paragraph{func\_generation.txt}
This prompt is the system prompt of all of the function generation process.
\begin{quote}
You are an expert in heuristic design. Your task is to complete the functions in the given heuristic structure to meet the following requirements:

1. Your design must fully satisfy the requirements specified in the provided function templates.

2. Your design must maintain the same parameters as the given function templates.

3. Your design should work correctly within the context of the heuristic, which will be provided below.

Your response must include the complete heuristic code with all necessary functions filled in.

Your response outputs Python code and nothing else. Format your code as a Python code string: "\texttt{\`}\texttt{\`}\texttt{\`}python ... \texttt{\`}\texttt{\`}\texttt{\`}".

\end{quote}

\paragraph{heubase\_common.txt}
This prompt is used to provide LLM with AM as well as the HeuBase.
\begin{quote}
Below is the heuristic database. 

You can directly call any of them. (Note that `edge\_index` is sized (2, num\_edge) for unweighted graph and is sized (3, num\_edge) for weighted graph where the third dimension is the length of the edge; `ini\_sol` must be valid)

Caution! Do not reimplement these functions—they are preloaded and conflicts will arise if duplicated. Simply call them by name with the required arguments.

\end{quote}

\paragraph{mutation.txt}
This prompt is the mutation prompt. the \{\} part will be fill in with the relative-structures.

\begin{quote}
\{exterior\_user\_generator\}\\

[Now Structure]

\{now\_structure\}\\

[Elitist Code]

\{elitist\_structure\}\\

[Improved code]

Please write a mutated structure based on the Now Structure. You should reflect on why the elitist code perform the best and take inspiration from it. Enclose your code with a Python fenced code block.
\end{quote}

\paragraph{pip\_search.txt}

This prompt is used to require the LLM to search for the relative libraries related that may help construct the heuristic. We adopt Lepton AI\footnote{\url{https://github.com/leptonai/search_with_lepton}} for online search.

\begin{quote}
We are solving a problem of {problem\_name}, {problem\_description}. You are required to search for some libraries for python that has a close relationship with this problem in topics or in details. Your output should follow this, act like a requirements.txt:

\texttt{\`}\texttt{\`}\texttt{\`}

[library\_name\_1] == [version\_number]

[library\_name\_2] == [version\_number]

\texttt{\`}\texttt{\`}\texttt{\`}
\end{quote}

\paragraph{prior\_knowledge.txt}
This prompt is used to provide LLM with KnoBase.
\begin{quote}
You may refer to these prior expert knowledge:

\{prior\_knowledge\}
\end{quote}

\paragraph{problem\_description.txt}
This prompt is used to provide LLM with problem description.

\begin{quote}
The problem description is as follows:

\{problem\_description\}
\end{quote}

\paragraph{system\_generator.txt}
This prompt is the system prompt of the whole process.
\begin{quote}
You are an expert in the domain of optimization heuristics. Your task is to design heuristics that can effectively solve optimization problems.

Your response outputs Python code and nothing else. Format your code as a Python code string: "\texttt{\`}\texttt{\`}\texttt{\`}python ... \texttt{\`}\texttt{\`}\texttt{\`}".
\end{quote}

\subsection{Prompt Format for Specific Problems}
\paragraph{description.txt}
This is different for every problem. Note that this is optional, because when facing a new problem, LLM needs to know the definition of the problem, while facing an old problem, there is no such need. For each problem, the description of the problem will be provided here. 

\paragraph{function\_signature.txt}
This is different for every problem. For each problem, a function signature will be provided to specify the required input and output format.
\begin{quote}
\texttt{\`}\texttt{\`}\texttt{\`}python

def heuristic (...parameters...) -> ...

    """

    Args:...

    Returns:...

    """
    
\texttt{\`}\texttt{\`}\texttt{\`}
\end{quote}

\paragraph{heubase.txt}
This is different for every problem. For each problem, this will provide LLM with Heubase and Adaptive Memory function's conclusive summary, so that LLM can know the function's usage without the need to read the actual code. The general structure is shown below.
\begin{quote}
\texttt{\`}\texttt{\`}\texttt{\`}python

def func\_name (...parameters...):

    """

    Usage:...

    Args:...

    Returns:...

    """

\texttt{\`}\texttt{\`}\texttt{\`}\\
\texttt{\`}\texttt{\`}\texttt{\`}python

def func\_name (...parameters...):

    """

    Usage:...

    Args:...

    Returns:...

    """

\texttt{\`}\texttt{\`}\texttt{\`}\\
... ...

\end{quote}

\paragraph{knobase.txt}
This is different for every problem and is written by LLM Researcher.

\section{Generated Codes \& Detailed Comparison}
\label{sec:generatedCodes}

\subsection{Generated Codes Comparison with other LHHs}
\label{sec:generatedcompare}
\begin{figure*}
    \centering
    \includegraphics[width=0.8\linewidth]{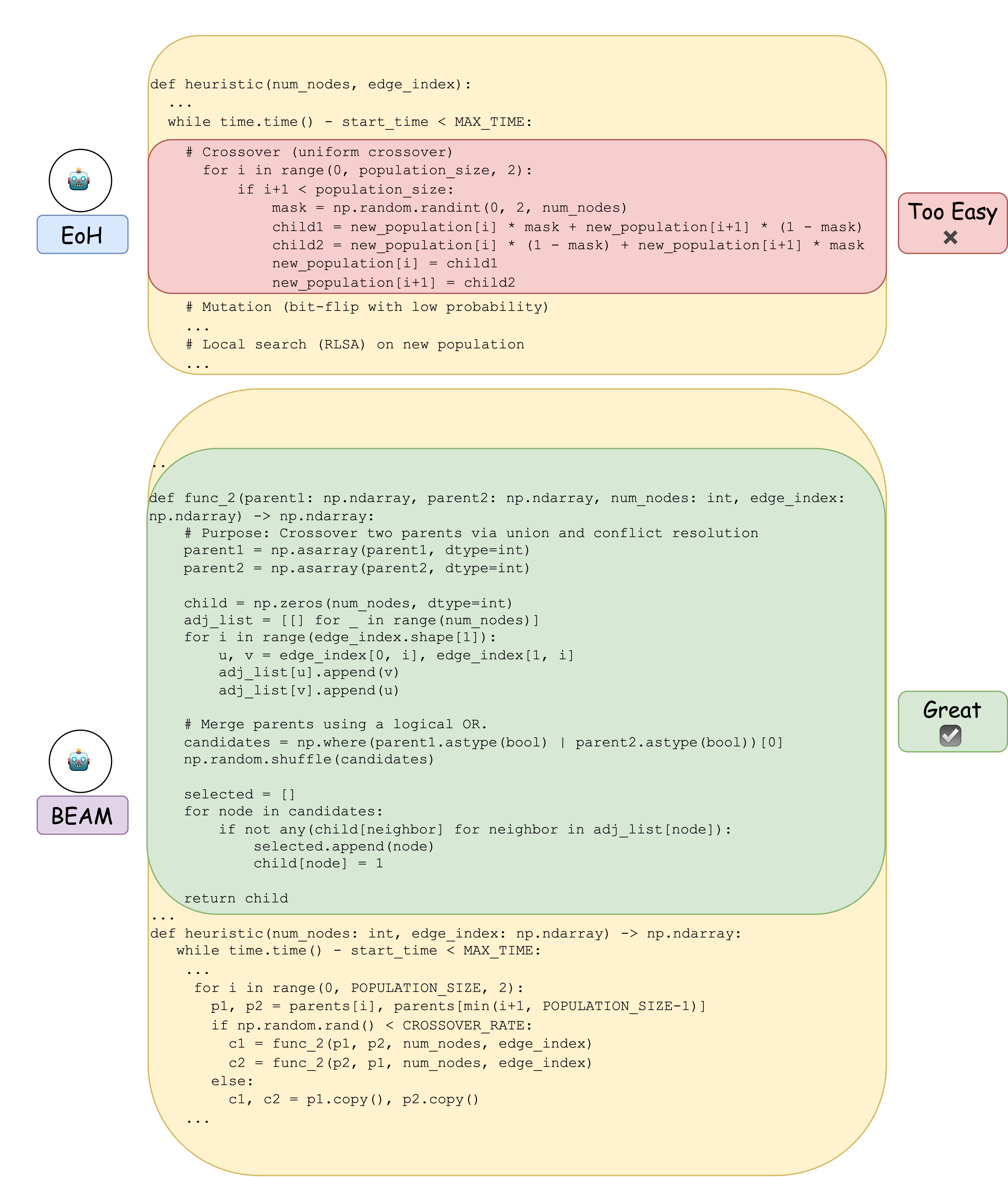}
    \caption{EoH vs. BEAM: MIS solver with RLSA.}
    \label{fig:codeComparison}
\end{figure*}

To further exemplify the algorithmic complexity differences mentioned above, we compare the MIS solver generated by BEAM and EoH. From Table~\ref{tab:CombinedTable}, we can find that the performance gap between BEAM and EoH widens significantly on harder instances (RB 800-1200), as shown in Table~\ref{tab:CombinedTable}. This divergence stems from EoH's oversimplified crossover mechanism - it relies exclusively on uniform crossover (See Fig.~\ref{fig:codeComparison}). While this simplicity may leave more computational budget for RLSA local search on smaller instances (RB-Small), it fundamentally limits EoH’s ability to escape local optima on larger, more challenging instances (RB-Large). In contrast, BEAM’s more sophisticated evolutionary framework enables stronger capabilities, leading to consistently better performance as problem difficulty increases.

\begin{figure*}
    \centering
    \includegraphics[width=1\linewidth]{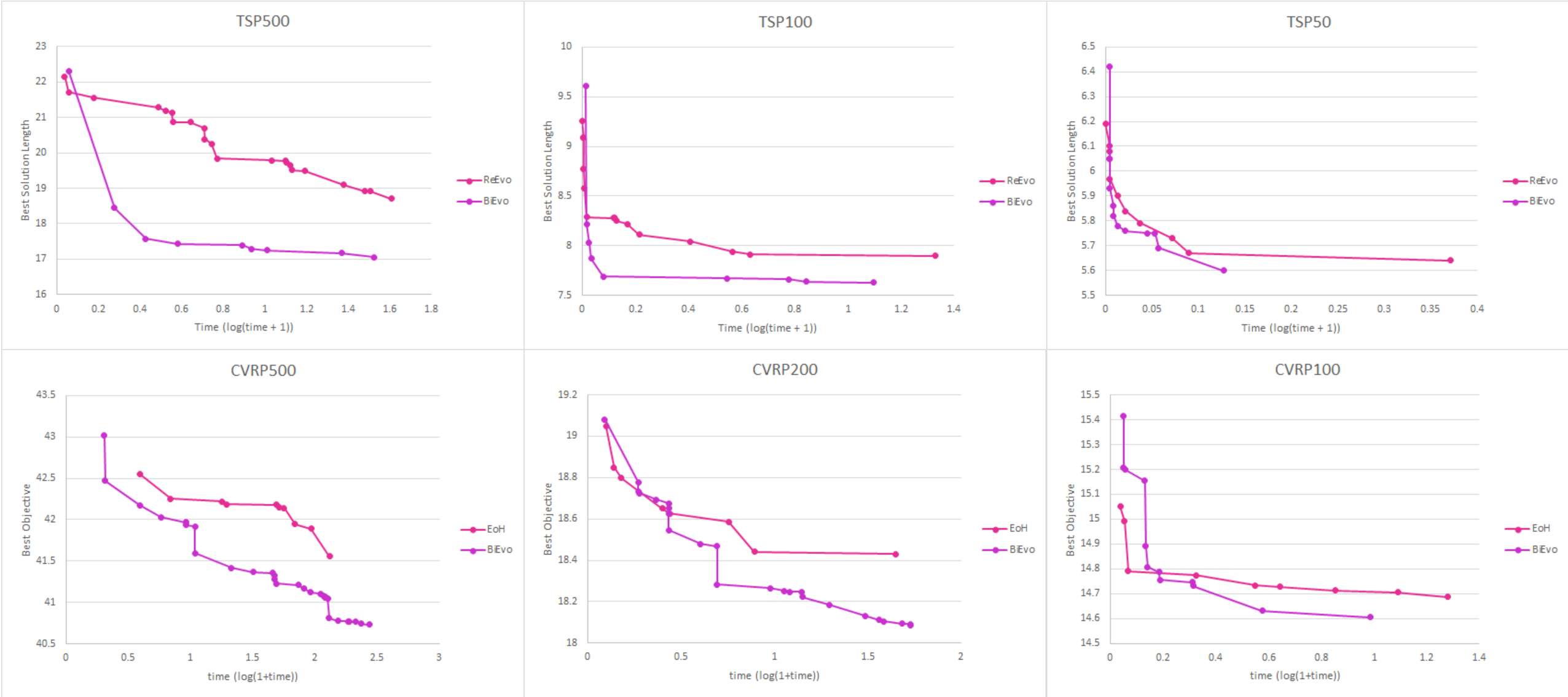}
    \caption{Generated Algorithm Iteration Curve.}
    \label{fig:algorithmshoulian}
\end{figure*}

For TSP and CVRP, the conclusions are similar, with Fig.~\ref{fig:algorithmshoulian} illustrating the iteration curve of the generated algorithms (the curve is averaged over 5 generated algorithms).

\subsection{Generated Codes Introduction}
\label{sec:codeintro}

\paragraph{TSP - Traditional Benchmark}
The best algorithm (Lisiting~\ref{lst:gls}) features a two-stage computation process that first calculates edge utilities from normalized distances and then transforms them into penalties using configurable hyperparameters (ALPHA, BETA) and non-linear scaling. This algorithm is generated within a fixed budget, so it isn't necessarily the best.

\paragraph{BPP - Traditional Benchmark}
The best algorithm (Listing~\ref{lst:bpp}) implements a three-phase, time-aware bin selection strategy for Bin Packing. It blends capacity-based and fit-based scoring, leverages a fast exact-fit shortcut, aggressively avoids overfilling bins via a lookahead penalty, and refines selections under tight time constraints. This algorithm is generated within a fixed budget, so it isn't necessarily the best.

Despite its good performance compared with its counterparts, we must note that BEAM tends to overcomplicate solutions for simple objectives - a tendency clearly reflected in code length. While EoH-generated solutions typically maintain concise implementations under 20 lines, BEAM's output often exhibits unnecessary complexity.

\paragraph{CAF - Traditional Benchmark}
The best algorithm (Listing~\ref{lst:caf}) dynamically adjusts exploration and exploitation priorities based on optimization progress, which is jointly characterized by budget consumption and solution quality. By blending standard and phase-aware Expected Improvement (EI) and scaling cost penalties according to phase, the method ensures robust and adaptive decision-making. The utility function further amplifies this adaptivity through exponentiation and diminishing sensitivity to cost over time.

\paragraph{MIS with RLSA}
The best algorithm (Listing~\ref{lst:rlsa}) is overall a standard memetic algorithm, standing out by evolving entirely within the feasible independent set space—thanks to heuristic initialization, conflict-free crossover, and RLSA-driven local search—which accelerates convergence and ensures consistently valid, high-quality solutions. A comparison with EoH is given in Fig.~\ref{fig:codeComparison}.

\paragraph{MIS with KaHIP \& ARW}
The best algorithm (Listing~\ref{lst:kahip}) utilizes KaHIP~\cite{KaHIP} in the initialization stage by isolating populations per partition and enabling occasional inter-block exchange, which is the best usage of KaHIP BEAM has found. 

Another interesting finding is that this algorithm, which outperforms KaMIS, relies on a simple uniform crossover—unlike KaMIS~\cite{KaMIS}, which uses KaHIP specifically in the crossover stage. To further investigate, we manually replaced the uniform crossover with KaHIP-based crossover, mimicking KaMIS’s approach. Surprisingly, this change led to \textit{worse performance}, suggesting that utilizing KaHIP in the crossover stage doesn't necessarily boost the performance.

\paragraph{CVRP with Split \& LS}
The best algorithm (Listing~\ref{lst:cvrp}) combines 4 initialization strategies to create a diverse initial population of solutions. It features an adaptive perturbation mechanism~\cite{Perturb} that employs 4 different mutation strategies with problem-size-dependent intensity for effective exploration. The implementation also incorporates periodic intensification to refine good solutions. 

It's worth noting that this codes serve as a good example of MCTS's strength since from the comment in \texttt{func\_1} we can know that LLM only plans to realize two mutation strategies (swap and reverse) in exterior structure evolution. When realizing the function in the interior layer, LLM expands the strategy sets with two more complicated strategies (shift and scramble). Similarly, \texttt{func\_2} provides a correct evaluation function, which is also meaningful since we find that single-layered EoH may even output the wrong evaluation function.

\paragraph{TSP with EDM}
The best algorithm (Listing~\ref{lst:edm}) implements a well-designed ACO framework. Its core strength lies in the balanced integration of pheromone-guided exploration and adaptive 2-opt refinement, enabling robust performance across diverse problem scales. The algorithm preserves solution quality via elite-preservation mechanisms and on-demand local optimization

\paragraph{BBOB}
The best algorithm (Listing~\ref{lst:bbob}) intelligently combines differential evolution, CMA-ES, and PSO in a staged optimization framework. Its key advantage lies in the dynamic allocation of computational budgets to each method based on their complementary strengths - DE for broad exploration, CMA-ES for precise local refinement, and PSO for final polishing. It automatically adjusts critical parameters during optimization, delivering robust performance across diverse continuous optimization landscapes. A smart restart mechanism further enhances solution quality.

\subsection{Codes generated by BEAM}

\begin{lstlisting}[language=Python, label={lst:gls}, caption=BEAM-generated penalty function for TSP-GLS.]
ALPHA = 0.3
BETA = 1.5
INIT_PENALTY = 1

import numpy as np

def heuristic(distance_matrix: np.ndarray) -> np.ndarray:
    # The `heuristic` function takes as input a distance matrix, and returns prior indicators of how bad it is to include each edge in a solution. The return is of the same shape as the input.
    
    # Step 1: Calculate initial edge utilities based on distance matrix
    edge_utilities = func_1(distance_matrix)
    
    # Step 2: Compute penalties based on edge utilities and current penalties
    penalties = func_2(edge_utilities, ALPHA, BETA, INIT_PENALTY)
    
    return penalties

def func_1(distance_matrix: np.ndarray) -> np.ndarray:
    # Purpose: Calculate initial edge utilities based on distance matrix (higher distance = higher utility)
    # Normalize the distance matrix to [0,1] range for utility calculation
    max_dist = np.max(distance_matrix)
    if max_dist > 0:
        normalized_dist = distance_matrix / max_dist
    else:
        normalized_dist = distance_matrix
    return normalized_dist

def func_2(edge_utilities: np.ndarray, alpha: float, beta: float, init_penalty: float) -> np.ndarray:
    # Purpose: Compute penalties for edges based on their utilities and hyperparameters
    # Calculate penalty adjustment factor
    utility_factor = alpha * edge_utilities
    # Apply non-linear transformation using beta exponent
    penalty_adjustment = np.power(utility_factor, beta)
    # Compute final penalties by scaling with initial penalty
    penalties = init_penalty * penalty_adjustment
    return penalties
\end{lstlisting}

\begin{lstlisting}[language=Python, label={lst:bpp}, caption=BEAM-generated BPP priority function.]
MAX_TIME = 2
FITNESS_WEIGHT = 0.5514079756555896
CAPACITY_WEIGHT = 0.45840214751046593
STABILITY_WEIGHT = 0.10166942364627612
FILL_THRESHOLD = 0.7993217820159937
TIME_CHECK_INTERVAL = 0.0021170327307504515
LOOKAHEAD_PENALTY = 0.4432174448805559
CORE_PHASE_RATIO = 0.6278250358826285

import numpy as np
import time

def heuristic(item: float, bins_remain_cap: np.ndarray) -> np.ndarray:
    """
    Hybrid heuristic combining the best elements from both approaches:
    1. Maintains three-phase structure but with improved time allocation
    2. Uses capacity-based scoring inspired by elitist code
    3. Incorporates more aggressive fill threshold from elitist version
    4. Optimized time checks and weight distribution
    """
    start_time = time.time()
    
    # Phase 0: Fast exact fit check (immediate return if found)
    if time.time() - start_time > MAX_TIME:
        return np.zeros_like(bins_remain_cap)
    
    exact_fit_mask = (bins_remain_cap == item)
    if np.any(exact_fit_mask):
        return np.where(exact_fit_mask, np.inf, -np.inf)
    
    # Initialize scores with capacity validation
    valid_bins = bins_remain_cap >= item
    priority_scores = np.where(valid_bins, 0.0, -np.inf)
    
    # Phase 1: Core calculations (time-constrained)
    if time.time() - start_time < MAX_TIME * CORE_PHASE_RATIO:
        # Parallel score calculations with frequent time checks
        capacity_scores = func_1(bins_remain_cap, item)
        if time.time() - start_time > MAX_TIME:
            return np.where(valid_bins, capacity_scores, -np.inf)
            
        fitness_scores = func_2(bins_remain_cap, item)
        if time.time() - start_time > MAX_TIME:
            combined = FITNESS_WEIGHT * fitness_scores + CAPACITY_WEIGHT * capacity_scores
            return np.where(valid_bins, combined, -np.inf)
        
        # Combine scores with weights
        priority_scores = np.where(
            valid_bins,
            FITNESS_WEIGHT * fitness_scores + CAPACITY_WEIGHT * capacity_scores,
            -np.inf
        )
        
        # Apply adaptive penalty (stronger than original)
        nearly_full = (bins_remain_cap / (bins_remain_cap + item)) > FILL_THRESHOLD
        priority_scores[nearly_full] *= LOOKAHEAD_PENALTY
    
    # Phase 2: Strategic refinement (if time permits)
    if time.time() - start_time < MAX_TIME * 0.9:
        priority_scores = func_3(priority_scores, bins_remain_cap, item)
    
    return priority_scores

def func_1(bins_remain_cap: np.ndarray, item: float) -> np.ndarray:
    # Purpose: Calculate capacity-based priority (normalized remaining capacity)
    max_cap = np.max(bins_remain_cap)
    if max_cap == 0:
        return np.zeros_like(bins_remain_cap)
    return bins_remain_cap / max_cap

def func_2(bins_remain_cap: np.ndarray, item: float) -> np.ndarray:
    # Purpose: Calculate fit-based priority (1 - abs(remaining_cap - item)/item)
    if item == 0:
        return np.zeros_like(bins_remain_cap)
    fit_quality = 1 - np.abs(bins_remain_cap - item) / item
    return np.maximum(0, fit_quality)  # Ensure non-negative scores

def func_3(priority_scores: np.ndarray, bins_remain_cap: np.ndarray, item: float) -> np.ndarray:
    # Purpose: Apply look-ahead adjustment considering potential future items
    # Penalize bins that would leave too little remaining capacity for typical future items
    remaining_after_packing = bins_remain_cap - item
    avg_item_size = item  # Using current item as estimate
    future_fit_penalty = np.where(
        remaining_after_packing < avg_item_size * 0.5,
        0.7,  # Strong penalty if unlikely to fit another item
        1.0    # No penalty if likely to fit another item
    )
    return priority_scores * future_fit_penalty

\end{lstlisting}

\begin{lstlisting}[language=Python, label={lst:caf}, caption=BEAM-generated CAF.]
from heubase.caf import EI  # This function is designed by BEAM
from heubase.caf import phase_aware_EI # This function is designed by BEAM
from heubase.caf import phase_aware_cost_scaling # This function is designed by BEAM

MAX_TIME = 2
COST_DISCOUNT_FACTOR = 0.4506552556778862
IMPROVEMENT_BOOST = 3.2386305441223917
EXPLORATION_FACTOR = 0.3038707156753378
UTILITY_EXPONENT = 1.8344236311494773
PHASE_SMOOTHING = 0.6121269499903754
COST_PENALTY = 0.12516595027522082

import torch
import time

def heuristic(train_x: torch.Tensor, 
              train_y: torch.Tensor, 
              best_x: torch.Tensor, 
              best_y: int, 
              test_x: torch.Tensor, 
              mean_test_y: torch.Tensor, 
              std_test_y: torch.Tensor, 
              cost_test_y: torch.Tensor, 
              budget_used: int, 
              budget_total: int
              ) -> torch.Tensor:
    """
    Optimized heuristic combining:
    1. Dynamic phase calculation with budget and quality awareness
    2. Balanced exploration-exploitation tradeoff
    3. Phase-aware non-linear utility combination
    4. Strict time constraints with frequent checks
    5. Cost penalty that increases with phase
    """
    start_time = time.time()
    
    # Calculate optimization phase considering both budget and solution quality
    phase = func_1(budget_used, budget_total, best_y, train_y, PHASE_SMOOTHING)
    
    if time.time() - start_time > MAX_TIME:
        return torch.zeros_like(mean_test_y)
    
    # Get both base and boosted EI values
    base_ei = EI(mean_test_y, std_test_y, best_y)
    boosted_ei = phase_aware_EI(
        mean=mean_test_y,
        std=std_test_y,
        best_y=best_y,
        phase=phase,
        improvement_boost=IMPROVEMENT_BOOST
    )
    
    if time.time() - start_time > MAX_TIME:
        return torch.zeros_like(mean_test_y)
    
    # Dynamic EI blending based on phase
    exploration_weight = EXPLORATION_FACTOR * (1 - phase)
    ei_values = (1 - exploration_weight) * boosted_ei + exploration_weight * base_ei
    
    # Get phase-sensitive cost scaling with additional penalty
    scaled_costs = phase_aware_cost_scaling(
        cost=cost_test_y,
        budget_used=budget_used,
        budget_total=budget_total,
        phase=phase,
        cost_discount_factor=COST_DISCOUNT_FACTOR
    ) * (1 + COST_PENALTY * phase)
    
    if time.time() - start_time > MAX_TIME:
        return torch.zeros_like(mean_test_y)
    
    # Phase-adaptive utility combination with exponent control
    utility = func_2(ei_values, scaled_costs, phase, UTILITY_EXPONENT)
    
    return utility

def func_1(budget_used: int, budget_total: int, best_y: float, train_y: torch.Tensor, smoothing: float) -> float:
    """
    Purpose:
    Calculate comprehensive optimization phase considering:
    1. Budget consumption ratio (linear)
    2. Solution quality improvement (non-linear)
    3. Smooth transitions between phases
    Returns normalized phase value [0,1]
    """
    # Budget-based phase component
    budget_phase = min(budget_used / budget_total, 1.0)
    
    # Quality-based phase component (normalized improvement)
    min_y = train_y.min().item()
    max_y = train_y.max().item()
    if max_y != min_y:
        quality_phase = (best_y - min_y) / (max_y - min_y)
    else:
        quality_phase = 0.0
    
    # Combined phase with smoothing
    combined_phase = smoothing * budget_phase + (1 - smoothing) * quality_phase
    return min(max(combined_phase, 0.0), 1.0)

def func_2(ei_values: torch.Tensor, scaled_costs: torch.Tensor, phase: float, exponent: float) -> torch.Tensor:
    """
    Purpose:
    Advanced utility function that:
    1. Uses exponential scaling for non-linearity
    2. Adapts cost sensitivity based on phase
    3. Maintains numerical stability
    4. Incorporates diminishing returns on high-cost evaluations
    """
    # Phase-adaptive cost sensitivity
    cost_sensitivity = 1.0 - 0.5 * phase  # Reduces cost sensitivity as optimization progresses
    
    # Numerically stable utility calculation with exponent
    eps = 1e-8
    utility = (ei_values + eps).pow(exponent) / (scaled_costs + eps).pow(cost_sensitivity)
    
    return utility

\end{lstlisting}

\begin{lstlisting}[language=Python, label={lst:rlsa}, caption=BEAM-generated MIS solver with RLSA.]
from rlsa import rlsa
import torch
from torch import Tensor
import numpy as np
import time

MAX_TIME = 120 # Manullay set by us
RLSA_ITERATIONS = 150 # Manullay set by us
POPULATION_SIZE = 47
MUTATION_RATE = 0.08673770360907226
CROSSOVER_RATE = 0.8442911280296675
TOURNAMENT_SIZE = 5


def func_1(num_nodes: int, edge_index: np.ndarray) -> np.ndarray:
    # Purpose: Generate a random valid initial solution (independent set)
    solution = np.zeros(num_nodes, dtype=int)
    adj_list = [[] for _ in range(num_nodes)]
    for i in range(edge_index.shape[1]):
        u, v = edge_index[0, i], edge_index[1, i]
        adj_list[u].append(v)
        adj_list[v].append(u)
    nodes = np.random.permutation(num_nodes)
    for node in nodes:
        if not any(solution[neighbor] for neighbor in adj_list[node]):
            solution[node] = 1
    return solution

def func_2(parent1: np.ndarray, parent2: np.ndarray, num_nodes: int, edge_index: np.ndarray) -> np.ndarray:
    # Purpose: Crossover two parents via union and conflict resolution
    parent1 = np.asarray(parent1, dtype=int)
    parent2 = np.asarray(parent2, dtype=int)
    
    child = np.zeros(num_nodes, dtype=int)
    adj_list = [[] for _ in range(num_nodes)]
    for i in range(edge_index.shape[1]):
        u, v = edge_index[0, i], edge_index[1, i]
        adj_list[u].append(v)
        adj_list[v].append(u)
    
    # Merge parents using a logical OR.
    candidates = np.where(parent1.astype(bool) | parent2.astype(bool))[0]
    np.random.shuffle(candidates)
    
    selected = []
    for node in candidates:
        if not any(child[neighbor] for neighbor in adj_list[node]):
            selected.append(node)
            child[node] = 1
    
    return child

def func_3(solution: np.ndarray, num_nodes: int, edge_index: np.ndarray, mutation_rate: float) -> np.ndarray:
    # Purpose: Mutate by flipping nodes while maintaining validity
    mutated = np.asarray(solution, dtype=int).copy()
    adj_list = [[] for _ in range(num_nodes)]
    for i in range(edge_index.shape[1]):
        u, v = edge_index[0, i], edge_index[1, i]
        adj_list[u].append(v)
        adj_list[v].append(u)
    
    for node in range(num_nodes):
        if np.random.rand() < mutation_rate:
            if mutated[node] == 1:
                mutated[node] = 0
            else:
                if all(mutated[neighbor] == 0 for neighbor in adj_list[node]):
                    mutated[node] = 1
    return mutated

def func_4(graph, x: Tensor, penalty_coeff: float) -> Tuple[Tensor, Tensor]:
    x_uq = x.unsqueeze(1)
    energy_term1 = torch.sum(x, dim=1)
    energy_term2 = torch.sum((torch.matmul(x_uq, graph) * x_uq).squeeze(1), 1)
    energy = -energy_term1 + penalty_coeff * energy_term2
    grad_term1 = torch.ones_like(x)
    grad_term2 = penalty_coeff * torch.matmul(graph, x.unsqueeze(-1)).squeeze(-1)
    grad = -grad_term1 + grad_term2
    return energy, grad

def heuristic(num_nodes: int, edge_index: np.ndarray) -> np.ndarray:
    start_time = time.time()
    best_solution = np.zeros(num_nodes, dtype=int)
    best_fitness = 0
    
    # Initialize population
    population = [func_1(num_nodes, edge_index) for _ in range(POPULATION_SIZE)]
    fitness = [ind.sum() for ind in population]
    best_idx = np.argmax(fitness)
    best_solution, best_fitness = population[best_idx].copy(), fitness[best_idx]
    print(f"Initial best: {best_fitness}")
    
    while time.time() - start_time < MAX_TIME:
        # Parent selection (tournament)
        parents = []
        for _ in range(POPULATION_SIZE):
            candidates = np.random.choice(POPULATION_SIZE, TOURNAMENT_SIZE, replace=False)
            best = candidates[np.argmax([fitness[c] for c in candidates])]
            parents.append(population[best])
        
        # Crossover and mutation
        offspring = []
        for i in range(0, POPULATION_SIZE, 2):
            p1, p2 = parents[i], parents[min(i+1, POPULATION_SIZE-1)]
            if np.random.rand() < CROSSOVER_RATE:
                c1 = func_2(p1, p2, num_nodes, edge_index)
                c2 = func_2(p2, p1, num_nodes, edge_index)
            else:
                c1, c2 = p1.copy(), p2.copy()
            c1 = func_3(c1, num_nodes, edge_index, MUTATION_RATE)
            c2 = func_3(c2, num_nodes, edge_index, MUTATION_RATE)
            c1 = rlsa(num_nodes, edge_index, c1, RLSA_ITERATIONS, True, func_4)
            c2 = rlsa(num_nodes, edge_index, c2, RLSA_ITERATIONS, True, func_4)
            offspring.extend([c1, c2])
        
        # Evaluate offspring
        offspring_fitness = [c.sum() for c in offspring]
        combined_pop = population + offspring
        combined_fit = fitness + offspring_fitness
        
        # Survival selection
        sorted_idx = np.argsort(combined_fit)[::-1][:POPULATION_SIZE]
        population = [combined_pop[i] for i in sorted_idx]
        fitness = [combined_fit[i] for i in sorted_idx]
        
        # Update best
        current_best = np.max(fitness)
        if current_best > best_fitness:
            best_fitness = current_best
            best_solution = population[np.argmax(fitness)].copy()
        print(f"Best after iteration: {current_best}")
        
        # Time check
        if time.time() - start_time >= MAX_TIME:
            break
    
    return best_solution

\end{lstlisting}

\begin{lstlisting}[language=Python, label={lst:kahip}, caption=BEAM-generated MIS solver with KaHIP \& ARW.]
import numpy as np
import time
import random
from heubase.kahip import node_separator
from heubase.arw import arw

MAX_TIME = 60
POPULATION_SIZE = 5
BLOCK_SIZE = 4
IMBALANCE = 0.1
SEED = 42
MUTATION_RATIO = 0.05
EXCHANGE_RATE = 0.2
# Calibration doesn't offer an increase in performance so we adopt the initial setting.

def func_1(xadj: np.ndarray, adjncy: np.ndarray, solution: np.ndarray) -> np.ndarray:
    num_nodes = len(solution)
    for i in range(num_nodes):
        if solution[i] == 1:
            for j in range(xadj[i], xadj[i+1]):
                neighbor = adjncy[j]
                if solution[neighbor] == 1:
                    solution[neighbor] = 0
    return solution

def func_2(parent1: np.ndarray, parent2: np.ndarray, xadj: np.ndarray, adjncy: np.ndarray) -> tuple[np.ndarray, np.ndarray]:
    # Purpose: Uniform crossover with mask generation
    mask = np.random.randint(0, 2, parent1.size).astype(bool)
    child1 = np.where(mask, parent1, parent2)
    child2 = np.where(mask, parent2, parent1)
    return child1, child2

def heuristic(num_nodes, edge_index, xadj, adjncy):
    start_time = time.time()
    best_solution = np.zeros(num_nodes, dtype=int)
    best_size = 0
    
    # Initialize BLOCK_SIZE separate populations based on graph partitions
    _, parts = node_separator(xadj, adjncy, num_blocks=BLOCK_SIZE, imbalance=IMBALANCE, seed=SEED, mode=0)
    populations = []
    for block in range(BLOCK_SIZE):
        population = []
        for _ in range(POPULATION_SIZE):
            ind = np.zeros(num_nodes, dtype=int)
            ind[parts == block] = np.random.randint(0, 2, size=np.sum(parts == block))
            population.append(func_1(xadj, adjncy, ind))
        populations.append(population)
    
    while time.time() - start_time < 60:
        # Evolve each population separately
        for block in range(BLOCK_SIZE):
            # Evaluate
            fitness = [np.sum(ind) for ind in populations[block]]
            best_idx = np.argmax(fitness)
            if fitness[best_idx] > best_size:
                best_solution = populations[block][best_idx].copy()
                best_size = fitness[best_idx]
                print(f"Best size: {best_size}")
            
            # Selection
            new_pop = []
            for _ in range(POPULATION_SIZE):
                a, b = random.sample(range(POPULATION_SIZE), 2)
                winner = a if fitness[a] > fitness[b] else b
                new_pop.append(populations[block][winner].copy())
            
            # Crossover
            for i in range(0, POPULATION_SIZE, 2):
                if i+1 >= POPULATION_SIZE:
                    break
                child1, child2 = func_2(new_pop[i], new_pop[i+1], xadj, adjncy)
                new_pop[i] = func_1(xadj, adjncy, child1)
                new_pop[i+1] = func_1(xadj, adjncy, child2)
            
            # Mutation
            for ind in new_pop:
                for _ in range(int(num_nodes * MUTATION_RATIO)):
                    idx = random.randint(0, num_nodes-1)
                    ind[idx] = 1 - ind[idx]
                func_1(xadj, adjncy, ind)
            
            # Local search on best
            new_pop[0] = arw(xadj, adjncy, new_pop[0])
            populations[block] = new_pop
        
        # Periodically exchange individuals between populations
        if random.random() < EXCHANGE_RATE:
            src, dest = random.sample(range(BLOCK_SIZE), 2)
            idx = random.randint(0, BLOCK_SIZE)
            populations[dest][idx] = populations[src][idx].copy()
    
    return best_solution
\end{lstlisting}

\begin{lstlisting}[language=Python, label={lst:cvrp}, caption=BEAM-generated CVRP solver with Split \& Local Search.]
from heubase.hgs import split
from heubase.hgs import LS_Valid
from heubase.hgs import LS_Invalid
from heubase.cvrp import sweep_init # This function is designed by BEAM
import numpy as np
from typing import Tuple

MAX_TIME = 300
INITIAL_POOL_SIZE = 6
PERTURB_STRENGTH = 0.15
LS_INTENSIFY_PROB = 0.5

import time
import random

def func_1(perm: np.ndarray) -> np.ndarray:
    # Purpose: Perform adaptive permutation perturbation using swap and reverse mutations
    n = len(perm)
    perturbed = perm.copy()
    
    # Determine perturbation strength based on problem size
    k = max(1, int(n * PERTURB_STRENGTH))
    
    # Randomly choose between different perturbation strategies
    strategy = np.random.choice(['swap', 'reverse', 'shift', 'scramble'])
    
    if strategy == 'swap':
        # Perform k random swaps
        for _ in range(k):
            i, j = np.random.choice(n, 2, replace=False)
            perturbed[i], perturbed[j] = perturbed[j], perturbed[i]
            
    elif strategy == 'reverse':
        # Reverse a random subsequence
        i = np.random.randint(0, n - k + 1)
        perturbed[i:i+k] = perturbed[i:i+k][::-1]
        
    elif strategy == 'shift':
        # Shift a random subsequence to a new position
        i = np.random.randint(0, n - k + 1)
        j = np.random.randint(0, n - k + 1)
        while abs(i - j) < k:  # Ensure meaningful shift
            j = np.random.randint(0, n - k + 1)
        segment = perturbed[i:i+k]
        remaining = np.delete(perturbed, slice(i, i+k))
        insert_pos = j if j < i else j - k
        perturbed = np.insert(remaining, insert_pos, segment)
        
    elif strategy == 'scramble':
        # Scramble a random subsequence
        i = np.random.randint(0, n - k + 1)
        segment = perturbed[i:i+k]
        np.random.shuffle(segment)
        perturbed[i:i+k] = segment
    
    return perturbed

def func_2(solution: np.ndarray, dist_matrix: np.ndarray) -> float:
    # Purpose: Calculate total route distance using precomputed distance matrix
    total_distance = 0.0
    prev_node = 0  # Start at depot
    
    for node in solution[1:]:  # Skip first depot (already accounted for in prev_node)
        total_distance += dist_matrix[prev_node, node]
        prev_node = node
    
    # Add return to depot from last node
    total_distance += dist_matrix[prev_node, 0]
    
    return total_distance

def heuristic(
    nbClients: int,
    nbVehicles: int,
    capacity: float,
    depot_coord: np.ndarray,
    nodes_coord: np.ndarray,
    demands: np.ndarray,
) -> np.ndarray:
    start_time = time.time()
    best_sol = None
    best_cost = float('inf')
    
    # Precompute distance matrix
    all_nodes = np.vstack([depot_coord, nodes_coord])
    dist_matrix = np.sqrt(((all_nodes[:, np.newaxis] - all_nodes)**2).sum(axis=2))
    
    # Generate diverse initial permutations
    initial_perms = []
    
    # 1. Nearest neighbor heuristic
    if time.time() - start_time < MAX_TIME and nbClients > 0:
        current = 0
        unvisited = set(range(1, nbClients+1))
        nn_perm = []
        while unvisited:
            nearest = min(unvisited, key=lambda x: dist_matrix[current, x])
            nn_perm.append(nearest)
            unvisited.remove(nearest)
            current = nearest
        initial_perms.append(np.array(nn_perm))
    
    # 2. Farthest insertion heuristic
    if time.time() - start_time < MAX_TIME and nbClients > 0:
        farthest = np.argmax(dist_matrix[0, 1:]) + 1
        ff_perm = [farthest]
        unvisited = set(range(1, nbClients+1)) - {farthest}
        while unvisited and time.time() - start_time < MAX_TIME:
            candidate = max(unvisited, key=lambda x: min(dist_matrix[x][y] for y in ff_perm))
            ff_perm.append(candidate)
            unvisited.remove(candidate)
        initial_perms.append(np.array(ff_perm))
        
    # 3. Sweep algorithm
    if time.time() - start_time < MAX_TIME and nbClients > 0:
        initial_perms.append(sweep_init(nodes_coord))
    
    # 4. Random permutations
    while len(initial_perms) < INITIAL_POOL_SIZE and time.time() - start_time < MAX_TIME:
        initial_perms.append(np.random.permutation(nbClients) + 1)
    
    # Evaluate initial solutions
    for perm in initial_perms:
        if time.time() - start_time >= MAX_TIME:
            break
        solution, valid = split(nbClients, nbVehicles, capacity, depot_coord, nodes_coord, demands, perm)
        if valid:
            improved_sol, improved_valid = LS_Valid(nbClients, nbVehicles, capacity, depot_coord, nodes_coord, demands, solution)
            current_cost = func_2(improved_sol, dist_matrix) if improved_valid else float('inf')
            if current_cost < best_cost:
                best_sol = improved_sol
                best_cost = current_cost
                print(f"Initial best: {best_cost}")
    
    # Main optimization loop
    while time.time() - start_time < MAX_TIME:
        if best_sol is None:  # Fallback initialization
            perm = np.random.permutation(nbClients) + 1
            solution, valid = split(nbClients, nbVehicles, capacity, depot_coord, nodes_coord, demands, perm)
            if valid:
                best_sol, best_cost = solution, func_2(solution, dist_matrix)
            continue
        
        # Perturb current best permutation
        current_perm = best_sol[best_sol != 0].astype(int)
        perturbed_perm = func_1(current_perm)
        
        # Split and improve
        new_sol, new_valid = split(nbClients, nbVehicles, capacity, depot_coord, nodes_coord, demands, perturbed_perm)
        if new_valid:
            improved_sol, improved_valid = LS_Valid(nbClients, nbVehicles, capacity, depot_coord, nodes_coord, demands, new_sol)
        else:
            improved_sol, improved_valid = LS_Invalid(nbClients, nbVehicles, capacity, depot_coord, nodes_coord, demands, new_sol)
        
        # Evaluate and update
        if improved_valid:
            current_cost = func_2(improved_sol, dist_matrix)
            if current_cost < best_cost:
                best_sol = improved_sol
                best_cost = current_cost
                print(f"Iteration best: {best_cost}")
        
        # Periodic intensification
        if np.random.rand() < LS_INTENSIFY_PROB and best_sol is not None:
            intensified_sol, intens_valid = LS_Valid(nbClients, nbVehicles, capacity, depot_coord, nodes_coord, demands, best_sol)
            if intens_valid:
                intens_cost = func_2(intensified_sol, dist_matrix)
                if intens_cost < best_cost:
                    best_sol = intensified_sol
                    best_cost = intens_cost
                    print(f"Intensification best: {best_cost}")
    
    # Final validity check
    if best_sol is None:
        perm = initial_perms[0] if initial_perms else np.arange(1, nbClients+1)
        best_sol, _ = split(nbClients, nbVehicles, capacity, depot_coord, nodes_coord, demands, perm)
    
    return best_sol.astype(int) if best_sol is not None else np.array([0], dtype=int)
\end{lstlisting}

\begin{lstlisting}[language=Python, label={lst:edm}, caption=BEAM-generated TSP solver with EDM]
import numpy as np
from scipy.spatial import distance_matrix
from heubase.tsp_aco import compute_tsp_edge_heuristics
from heubase import two_opt_local_search # This function is designed by BEAM
NUM_ANTS = 100
EVAPORATION_RATE = 0.1
ALPHA = 1.0
BETA = 2.0
INIT_PHEROMONE = 0.1
Q = 100.0
ELITE_FACTOR = 2.0
MAX_TIME = 30

import numpy as np
import time

def heuristic(node_coor: np.ndarray) -> np.ndarray:
    n = node_coor.shape[0]
    dist_matrix = np.sqrt(((node_coor[:, None] - node_coor)**2).sum(axis=2))
    np.fill_diagonal(dist_matrix, 1.0)
    heuristic_matrix = compute_tsp_edge_heuristics(dist_matrix)
    pheromone = np.full((n, n), INIT_PHEROMONE)
    best_path = None
    best_length = np.inf
    start_time = time.time()
    
    while time.time() - start_time < MAX_TIME:
        paths = []
        lengths = []
        
        # Generate ant paths
        for _ in range(NUM_ANTS):
            current = np.random.randint(n)
            path = [current]
            unvisited = set(range(n)) - {current}
            
            while unvisited:
                current_node = path[-1]
                next_node = func_1(
                    current_node, list(unvisited),
                    pheromone[current_node], heuristic_matrix[current_node],
                    ALPHA, BETA
                )
                path.append(next_node)
                unvisited.remove(next_node)
            
            # Apply local optimization
            optimized_path = two_opt_local_search(np.array(path), dist_matrix)
            if len(np.unique(optimized_path)) == n:
                path = optimized_path
            
            # Calculate open TSP length
            length = dist_matrix[path[:-1], path[1:]].sum()
            paths.append(path)
            lengths.append(length)
            
            # Update best solution
            if length < best_length:
                best_path = np.array(path)
                best_length = length
        
        # Pheromone update
        pheromone = func_2(
            pheromone, paths, lengths, 
            best_path, best_length, 
            EVAPORATION_RATE, Q, ELITE_FACTOR
        )
        
        # print(f"Best length: {best_length}, Path: {best_path}")
    
    return best_path.astype(int) if best_path is not None else np.arange(n)

def func_1(current_node: int, unvisited: list, pheromone_row: np.ndarray, heuristic_row: np.ndarray, alpha: float, beta: float) -> int:
    # Purpose: Select next node using probabilistic rule (pheromone^alpha * heuristic^beta)
    unvisited = np.array(unvisited)
    pheromone = pheromone_row[unvisited]
    heuristic = heuristic_row[unvisited]
    
    probabilities = (pheromone ** alpha) * (heuristic ** beta)
    probabilities /= probabilities.sum()
    
    return np.random.choice(unvisited, p=probabilities)

def func_2(pheromone: np.ndarray, paths: list, lengths: list, best_path: np.ndarray, best_length: float, evaporation_rate: float, q: float, elite_factor: float) -> np.ndarray:
    # Purpose: Update pheromone with evaporation, ant deposits, and elite reinforcement
    # Evaporation
    pheromone *= (1 - evaporation_rate)
    
    # Ant deposits
    for path, length in zip(paths, lengths):
        delta = q / length
        for i in range(len(path)-1):
            u, v = path[i], path[i+1]
            pheromone[u, v] += delta
            pheromone[v, u] += delta
    
    # Elite reinforcement
    if best_path is not None:
        delta_elite = elite_factor * q / best_length
        for i in range(len(best_path)-1):
            u, v = best_path[i], best_path[i+1]
            pheromone[u, v] += delta_elite
            pheromone[v, u] += delta_elite
    
    return pheromone
\end{lstlisting}

\begin{lstlisting}[language=Python, label={lst:bbob}, caption=BEAM-generated BBOB.]
import numpy as np
from typing import Tuple, Callable
import cma # pip-installed with the help of LLM Researcher

BOUND = 5.12
P_DE = 0.6
P_CMA = 0.3
P_PSO = 0.1
POP_SIZE = 50
F_MIN = 0.4
F_MAX = 0.9
CR0 = 0.9
PSO_N = 10
W0 = 0.9
W1 = 0.4
C1 = 2.0
C2 = 2.0


def heuristic(problem_func: Callable, dimension: int, fopt: float, budget: int = 20000):
    bounds = (-BOUND, BOUND)
    evals = 0
    bud_de = int(budget * P_DE)
    bud_cma = int(budget * P_CMA)
    bud_pso = budget - bud_de - bud_cma

    pop = bounds[0] + (bounds[1]-bounds[0]) * func_2((POP_SIZE, dimension))
    fit = np.array([problem_func(ind) for ind in pop]); evals += POP_SIZE
    best_idx = np.argmin(fit)
    best_x, best_f = pop[best_idx].copy(), fit[best_idx]

    iters_de = bud_de // POP_SIZE
    for t in range(1, iters_de+1):
        F = F_MIN + (F_MAX - F_MIN)*(1 - t/iters_de)
        CR = CR0 * np.exp(-3*t/iters_de)
        for i in range(POP_SIZE):
            idxs = [j for j in range(POP_SIZE) if j!=i]
            a,b,c = pop[np.random.choice(idxs,3,replace=False)]
            # current-to-best/1
            mutant = pop[i] + F*(best_x-pop[i]) + F*(a-b)
            mutant = np.clip(mutant, bounds[0], bounds[1])
            mask = np.random.rand(dimension) < CR
            if not mask.any(): mask[np.random.randint(dimension)] = True
            trial = np.where(mask, mutant, pop[i])
            fv = problem_func(trial); evals+=1
            if fv < fit[i]:
                pop[i], fit[i] = trial, fv
                if fv < best_f:
                    best_x, best_f = trial.copy(), fv
        if t % 50 == 0:
            x2, f2 = func_1(best_x, problem_func, bounds)
            if f2 < best_f:
                best_x, best_f = x2.copy(), f2

    sigma0 = np.std(pop, axis=0).mean() + 1e-8
    es = cma.CMAEvolutionStrategy(best_x.tolist(), sigma0, {'popsize':20})
    while not es.stop() and evals < bud_de+bud_cma:
        X = es.ask()
        Fs = [problem_func(x) for x in X]; evals += len(X)
        es.tell(X, Fs)
    sol = np.array(es.result.xbest)
    fsol = problem_func(sol); evals+=1
    if fsol < best_f:
        best_x, best_f = sol.copy(), fsol

    pos = best_x + 0.1 * np.random.randn(PSO_N, dimension)
    vel = np.zeros_like(pos)
    pbest = pos.copy()
    pfit = np.array([problem_func(x) for x in pos]); evals+=PSO_N
    gbest, gfit = best_x.copy(), best_f
    for k in range(bud_pso//PSO_N):
        w = W0 + (W1-W0)*(k/(bud_pso//PSO_N))
        for i in range(PSO_N):
            r1, r2 = np.random.rand(dimension), np.random.rand(dimension)
            vel[i] = w*vel[i] + C1*r1*(pbest[i]-pos[i]) + C2*r2*(gbest-pos[i])
            pos[i] = np.clip(pos[i] + vel[i], bounds[0], bounds[1])
            fv = problem_func(pos[i]); evals+=1
            if fv < pfit[i]:
                pbest[i], pfit[i] = pos[i].copy(), fv
                if fv < gfit:
                    gbest, gfit = pos[i].copy(), fv
    best_x, best_f = gbest, gfit

    x3, f3 = func_1(best_x, problem_func, bounds, trials=20)
    if f3 < best_f:
        best_x, best_f = x3, f3

    gap = best_f - fopt
    return best_x.tolist(), round(best_f,6), round(gap,6)


def func_1(x: np.array, prob: Callable, bounds: float, trials: int = 10) -> float:
    n = len(x)
    H = np.random.choice([1, -1], size=(trials, n))
    cand = x + 1e-2 * H
    cand = np.clip(cand, bounds[0], bounds[1])
    vals = [prob(c) for c in cand]
    idx = np.argmin(vals)
    return cand[idx], vals[idx]

def func_2(shape: Tuple, mu: float = 0.3, iter: int = 5) -> Tuple:
    z = np.random.rand(*shape)
    for _ in range(iter):
        z = np.where(z < mu, z/mu, (1 - z)/(1 - mu))
    return z
\end{lstlisting}

\end{document}